\documentclass[default,iicol]{sn-jnl}
\usepackage{amsmath,amsfonts,amssymb}
\usepackage{algorithmic}
\usepackage{algorithm}
\usepackage{array}
\usepackage[table,xcdraw]{xcolor}
\usepackage{textcomp}
\usepackage{stfloats}
\usepackage{url}
\usepackage{verbatim}
\usepackage{graphicx}
\usepackage{cite}
\usepackage{mathrsfs}
\usepackage{epsfig}
\usepackage{enumerate}
\usepackage{cases}
\usepackage{multirow}
\usepackage{CJK}
\usepackage{subfigure}
\usepackage{threeparttable}
\usepackage{booktabs}
\usepackage{bm}
\usepackage{float}
\usepackage{ragged2e}
\usepackage{csquotes}
\usepackage{hyperref}
\usepackage{xspace}
\usepackage{overpic}
\usepackage{contour}
\usepackage{tablefootnote}
\usepackage{adjustbox}
\DeclareMathOperator*{\argmin}{argmin}
\newcommand{\tabincell}[2]{\begin{tabular}{@{}#1@{}}#2\end{tabular}}
\newcommand{\ie}{\textit{i.e.}}
\newcommand{\eg}{\textit{e.g.}}
\def\etal{\emph{et al.}}

\def\etal{\emph{et al.}}
\usepackage{color,soul}
\usepackage{xcolor}
\definecolor{bb}{rgb}{0.0, 0.0, 0.5}
\definecolor{Gray}{gray}{0.9}
\newcommand{\chong}[1]{\textcolor{black}{{#1}}}
\newcommand{\bihan}[1]{\textcolor{black}{{#1}}}

\begin{document}

\title[Article Title]{Single-Image Shadow Removal Using Deep Learning: A Comprehensive Survey}


\author[1]{\fnm{Lanqing} \sur{Guo}}\email{lanqing001@ntu.edu.sg}

\author[1]{\fnm{Chong} \sur{Wang}}\email{wang1711@ntu.edu.sg}

\author[1]{\fnm{Yufei} \sur{Wang}}\email{yufei001@ntu.edu.sg}

\author[1]{\fnm{Yi} \sur{Yu}}\email{yuyi0010@ntu.edu.sg}

\author[2]{\fnm{Siyu} \sur{Huang}}\email{siyuh@clemson.edu}

\author[3]{\fnm{Wenhan} \sur{Yang}}\email{yangwh@pcl.ac.cn}

\author[1]{\fnm{Alex C.} \sur{Kot}}\email{eackot@ntu.edu.sg}

\author*[1]{\fnm{Bihan} \sur{Wen}}\email{bihan.wen@ntu.edu.sg}

\affil[1]{\orgdiv{ROSE Lab}, \orgname{Nanyang Technological University}, \orgaddress{\street{50 Nanyang Ave}, \postcode{639798}, \country{Singapore}}}

\affil[2]{\orgdiv{Visual Computing Division, School of Computing}, \orgname{Clemson University}, \orgaddress{\street{SC 29631}, \city{Clemson}, \country{USA}}}

\affil[3]{\orgdiv{Pengcheng Laboratory}, \orgaddress{\street{No. 2, Xingke 1st Street}, \city{Shenzhen}, \postcode{518066}, \country{China}}}

\abstract{
Shadow removal aims at restoring the image content within shadow regions, pursuing a uniform distribution of illumination that is consistent between shadow and non-shadow regions. 
\bihan{Comparing to other image restoration tasks, there are two unique challenges in shadow removal:}
1) The patterns of shadows are arbitrary, varied, and often have highly complex trace structures, making ``trace-less'' image recovery difficult. 2) The degradation caused by shadows is spatially non-uniform, resulting in inconsistencies in illumination and color between shadow and non-shadow areas.
Recent developments in this field are primarily driven by deep learning-based solutions, employing a variety of learning strategies, network architectures, loss functions, and training data. Nevertheless, a thorough and insightful review of deep learning-based shadow removal techniques is still lacking.
In this paper, we are the first to provide a comprehensive survey to cover various aspects ranging from technical details to applications. 
We highlight the major advancements in deep learning-based single-image shadow removal methods, thoroughly review previous research across various categories, and provide insights into the historical progression of these developments. Additionally, we summarize performance comparisons both quantitatively and qualitatively. Beyond the technical aspects of shadow removal methods, we also explore potential future directions for this field.
The related repository is released at: \url{https://github.com/GuoLanqing/Awesome-Shadow-Removal}.}

\keywords{Shadow Removal, Low-Level Vision, Deep Learning, Image Enhancement, Computational Photography}

\maketitle

\section{Introduction}\label{sec:intro}
Shadowing is a natural occurrence observed when certain regions of a surface receive less illumination compared to their neighboring areas. This happens when an opaque object obstructs the direct path of light between the surface and the light source.
Shadows in images and videos hinder both the human perception~\cite{cucchiara2003detecting,nadimi2004physical} as well as many subsequent vision tasks, \eg, object detection, tracking, and semantic segmentation~\cite{jung2009efficient,zhang2018improving,han2023relating}.

\begin{figure*}
    \centering
    \includegraphics[width=15.5cm]{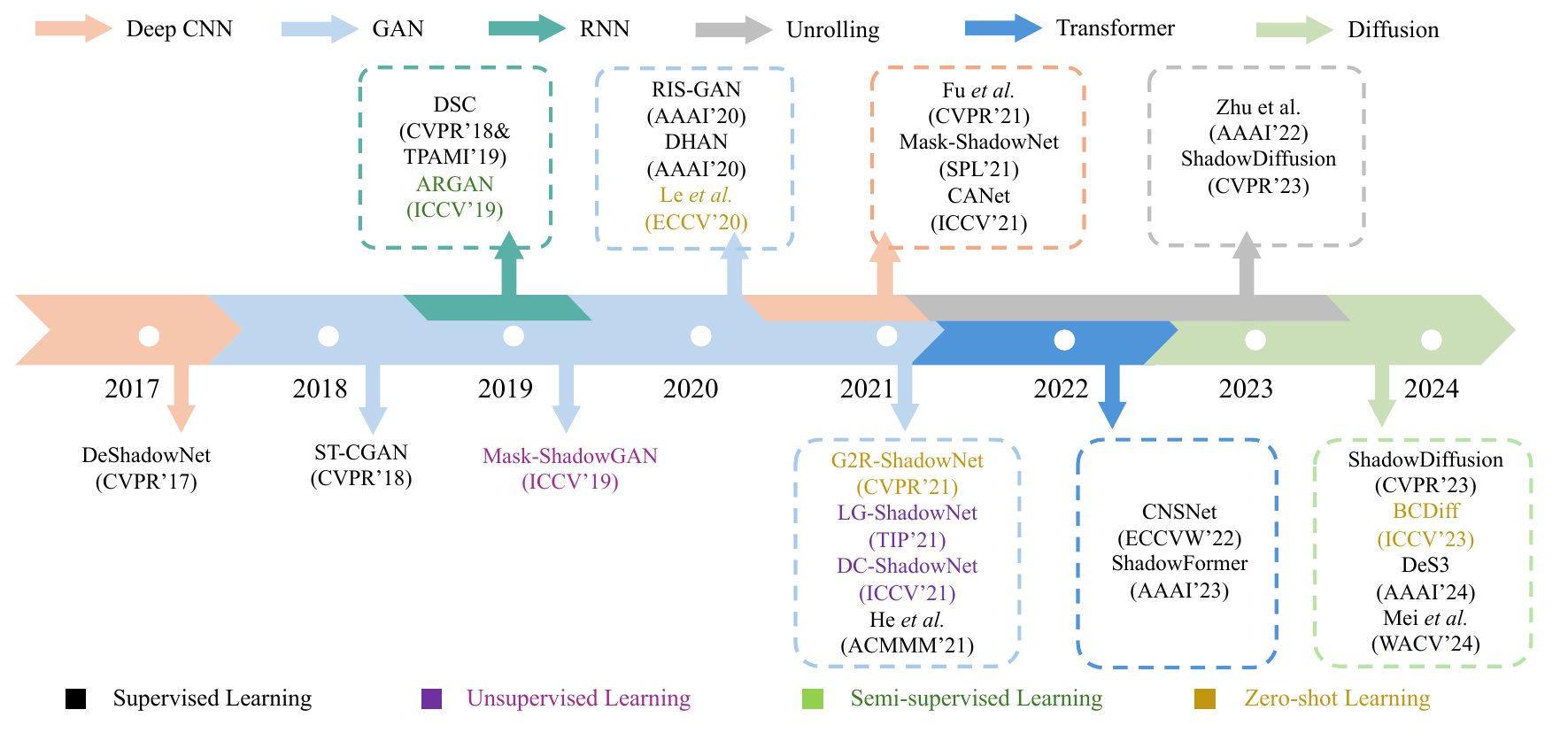}
    \caption{Milestones in deep learning-based single-image shadow removal methods include the exploration of various technologies over time, such as deep CNNs, GANs, RNNs, unrolling, transformers, and diffusion models.}
    \label{fig:milestone}
\end{figure*}

Shadow degradations manifest in diverse types and exhibit highly non-uniform properties, presenting unique challenges for shadow removal compared to other image restoration tasks.
Traditional shadow removal methods leverage hand-crafted prior information such as illumination~\cite{arbel2007texture,fredembach2006simple,salamati2011removing,zhang2015shadow,guo2012paired,vicente2014single}, gradient~\cite{finlayson2002removing,finlayson2005removal,newfinlayson2002removing,fredembach2005hamiltonian,fredembach2004fast,gryka2015learning}, and color~\cite{wu2005bayesian,shor2008shadow,xiao2013fast,xiao2013efficient,khan2015automatic} to restore illumination in shadowed areas.
Typically, some methods~\cite{finlayson2005removal,finlayson2009entropy} eliminate the image gradient along shadow edges and then reintegrate the modified gradient to produce a shadow-free image.
However, these approaches are based on ideal assumptions, resulting in noticeable shadow boundary artifacts in real-world scenarios.
They always require accurate and fine-grained shadow edges in the detection output as the strong prior for shadow removal, which is impractical in applications;
another group~\cite{arbel2007texture,fredembach2006simple,salamati2011removing,zhang2015shadow,guo2012paired,vicente2014single} treat shadow removal as a relighting problem, seeking a factor to enhance the brightness of shadow pixels. The main challenge is to determine different scale factors for umbra and penumbra shadows, along with addressing shadow boundary corrections due to non-uniform illumination degradation.


In recent years, deep learning-based shadow removal methods~\cite{qu2017deshadownet,ding2019argan,guo2022shadowdiffusion,jiang2023learning} have demonstrated superior performance, owing to the availability of extensive training data. 
Figure~\ref{fig:milestone} presents a concise overview of the milestones achieved by deep learning-based shadow removal methods in the past years.
In 2017, Qu~\etal~\cite{qu2017deshadownet} first proposed a multi-branch fusion framework as the pioneering work to adopt a deep convolutional neural networks (CNNs)~\cite{lecun1989handwritten,krizhevsky2012imagenet,venkatesan2017convolutional} to this field. 
Subsequently, starting from 2018, the advancement of Generative Adversarial Networks (GANs)~\cite{goodfellow2020generative} in the realm of low-level vision has led to widespread adoption of GAN-based networks in shadow removal tasks. These networks are not only effective in addressing amplified artifacts but also enable the development of unsupervised learning methods using unpaired shadow and shadow-free data.
During this period, some methods tried to inject constraints, \eg, physical illumination model and mask information, into networks to learn more effective features, by unrolling~\cite{zhu2022efficient,guo2022shadowdiffusion} or recurrent~\cite{ding2019argan} strategies.
In the most recent developments, novel techniques such as the transformer mechanism~\cite{vaswani2017attention,dosovitskiy2020ViT,liu2021swin,liang2021swinir} and diffusion model~\cite{sohl2015deep,ho2020denoising,kawar2022denoising,saharia2022image,lugmayr2022repaint} have shown remarkable advancements in shadow removal.
The transformer mechanism exploits long-range dependencies within contexts, facilitating more effective extraction of context information from non-shadow regions. Meanwhile, the diffusion model offers a potent diffusive generative prior for generating natural shadow-free images.




Despite the dominance of deep learning in shadow removal research, there is a lack of an in-depth and comprehensive survey on deep learning-based solutions.
The existing surveys for shadow removal~\cite{murali2016survey,tiwari2016survey,das2013review} only cover traditional methods \bihan{based on earlier publications, but omit the recent works via} deep learning approaches. 
In this paper, we are the first to review and summarize deep learning-based single-image shadow removal methods, aiming to offer a well-structured and comprehensive knowledge base to support and advance the field of shadow removal.
The rest of the paper is organized as follows.
Section~\ref{sec:intro} introduces the shadow formulation model and the problem definition for the shadow removal task.
Section~\ref{sec:formation} presents a detailed shadow formation model and outlines the various shadow categories, highlighting the different challenges associated with each.
Section~\ref{sec:deep} provides a detailed survey of deep learning based single-image shadow removal of different learning strategies, including supervised, semi-supervised, unsupervised, and zero-shot learning, as well as summarises the challenges of analysis of shadow removal.
Subsequently, Section~\ref{sec:technical} reviews the related technical aspects and explores solutions to the aforementioned challenges through various technical designs and the integration of priors, including network architectures, basic blocks, framework designs, loss functions, datasets, and evaluation metrics.
Section~\ref{sec:benchmark} summarizes the quantitative and qualitative comparisons of a number of single-image shadow removal methods among various benchmarks. 
Section~\ref{sec:application} presents the related shadow removal applications, including shadow generation and shadow-related attack in practice.
Section~\ref{sec:future} provides detailed discussions of the future research directions on generalized shadow removal and interactive shadow removal.
Finally, the paper is concluded in Section~\ref{sec:conclusion}.

\section{Analysis on Shadow Formation}\label{sec:formation}

\subsection{Shadow Formation Model}
According to the Retinex theory~\cite{land1977retinex}, the formation process of a shadow-free image $\mathbf{I}_{sf}$ with normal lightness can be formulated as
\begin{equation}
    \mathbf{I}_{sf} = \mathbf{L}_{sf}\circ \mathbf{R}
\end{equation}
where $\mathbf{L}$ denotes the illumination, $\mathbf{R}$ denotes the reflectance, and $\circ$ denotes element-wise multiplication.
Shadows occur when an occluder blocks light between the surface and the light source, causing partial light degradations.
Thus, the corresponding shadow image $\mathbf{I}_{s}$ with degraded illumination $\mathbf{L}_{s}$ can be formulated as
\begin{equation}\label{eq:shadow_formation}
    \mathbf{I}_{s} = \mathbf{L}_{s}\circ\mathbf{R} = \mathbf{A}\circ\mathbf{L}_{sf}\circ\mathbf{R} ,
\end{equation}
where $\mathbf{A}$ denotes the spatially variant shadow degradation degrees at each location.

\subsection{Shadow Category}

\begin{figure*}
    \centering
    \includegraphics[width=16cm]{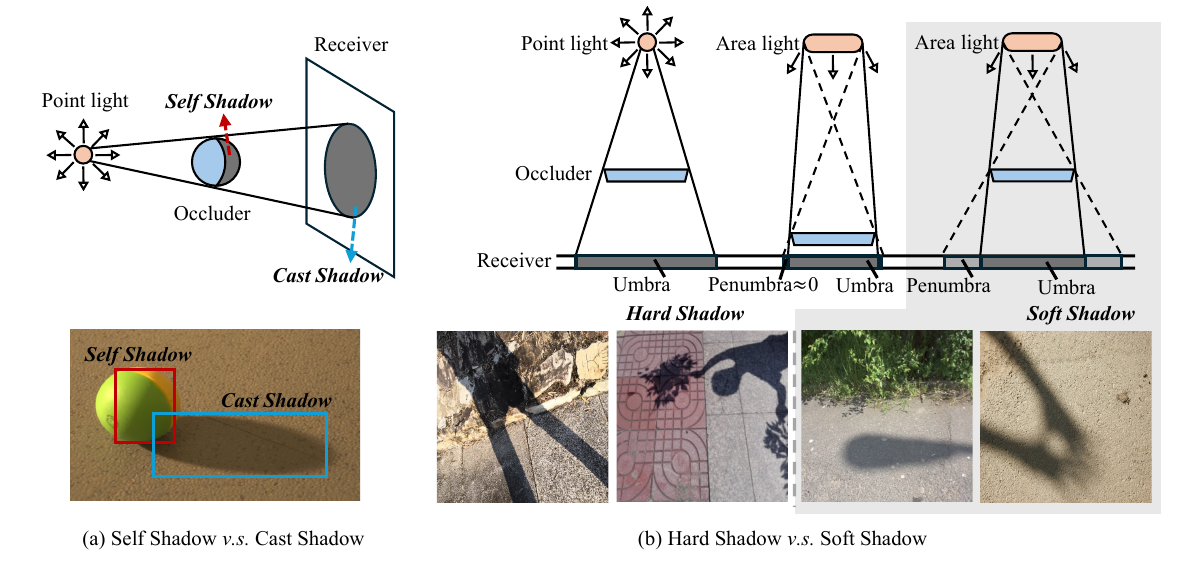}
    \caption{How a shadow is formed? (a) Shadows are then classified as \textit{cast shadows} if they belong to the background of the scene or as \textit{self shadows} if they are part of an occluder object.
    (b) The cast shadows can be further classified as \textit{hard shadows} and \textit{soft shadows}.
    A crisp edged one (hard shadow) formed by a point light source, a rather more fuzzy one (soft shadow) that is formed by the area light, and otherwise the occluder is very close to the receiver.}
    \label{fig:physic}
\end{figure*}

Shadow degradations have diverse types and highly non-uniform properties, which present unique challenges for shadow removal compared to other image restoration tasks.
Shadows can be classified  as \textit{self shadows} if they are part of an object or as \textit{cast shadows} if they belong to the background of the scene as shown in Figure~\ref{fig:physic}(a).
For a non-point source of light, a cast shadow is further sub-divided into \textit{soft and hard shadows} based on the illumination intensity (\ie, darkness) as shown in Figure~\ref{fig:physic}(b).
Specifically, shadows produced by the obstruction of a non-point light source by an object exhibit two distinct regions: the umbra and the penumbra. 
The umbra denotes the high-density region, located toward the inner shadow area, whereas the penumbra represents the low-density region, situated toward the outer shadow area.
Hard shadows exhibit umbra regions with a high density, often resulting in nearly obliterated surface texture. 
In contrast, soft shadows display penumbra regions towards the outer shadow area, with lower density, and their boundaries typically blend into the non-shadowed surroundings.
The distance between the occluder and receiver, as well as the type of light source, determines whether hard or soft shadows are produced, as depicted in Figure~\ref{fig:physic}(b).

Shadow degradations vary widely in type and exhibit highly non-uniform characteristics, requiring better generalization ability to handle varying cases in the real world. Different shadow types present distinct challenges: hard shadows with sharp boundaries can leave noticeable traces, while soft shadows and self-shadows, with ambiguous boundaries, are more difficult to detect and mask accurately.


\section{Deep Learning Based Shadow Removal}\label{sec:deep}

In this section, we first provide the problem definition of single-image shadow removal. Then, we review and discuss existing shadow removal methods according to the categories of general learning strategies. After that, we summarize the challenges and provide an analysis of existing methods.

\subsection{Problem Definition}
We begin by presenting the common formulation of the deep learning-based image shadow removal problem.
For a shadow image $\mathbf{I}_s \in \mathbb{R}^{H\times W\times 3}$ with width $W$ and height $H$, the process can be modeled as follows:
\begin{equation}
    \hat{\mathbf{I}}_{sf} = \mathcal{G}(\mathbf{I}_s; \theta)  ,
\end{equation}
where $\hat{\mathbf{I}}_{sf} \in \mathbb{R}^{H\times W\times 3}$ is the restored shadow-free result and $\mathcal{G}$ represents the shadow removal network with trainable parameters $\theta$. 
Different from other image restoration tasks, shadow removal is a partial corruption problem that requires identifying the shadow regions and restoring them.
To aid this process, an optional shadow mask $\mathbf{M} \in \mathbb{R}^{H\times W}$ can be used as auxiliary information, indicating shadow regions either manually annotated or detected by a pre-trained shadow detector.
The uniqueness property motivates numerous explorations into how shadow mask information can assist in shadow removal, facilitating the development of shadow mask-guided removal process as follows:
\begin{equation}
    \hat{\mathbf{I}}_{sf} = \mathcal{G}(\mathbf{I}_s, \mathbf{M}; \theta)\;.
\end{equation}



\subsection{General Learning Strategies}\label{sec:learning_strategy}

\begin{figure}
    \centering
    \includegraphics[width=7.5cm]{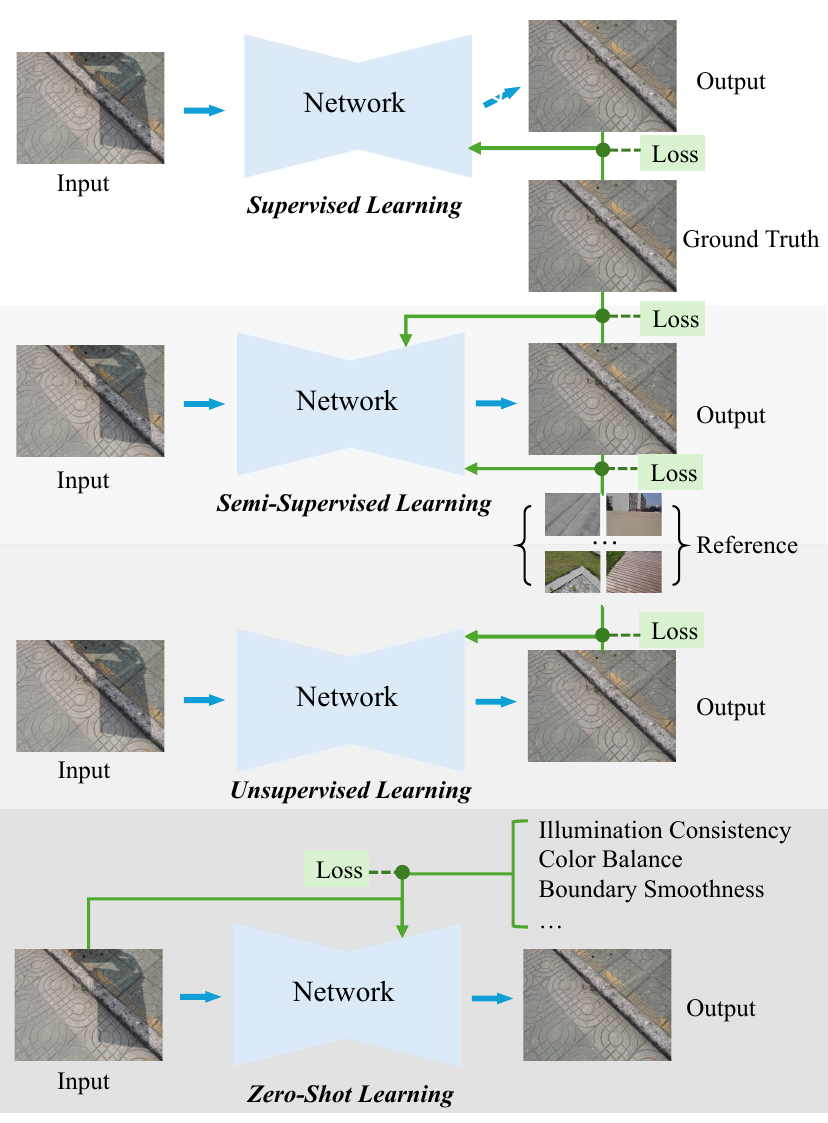}
    \caption{\bihan{Illustration of common deep learning strategies for single-image shadow removal.} Details refer to Section~\ref{sec:learning_strategy}.}
    \label{fig:learning_strategy}
\end{figure}

Based on different learning strategies, we generally categorize existing image shadow removal methods into supervised learning, unsupervised learning, zero-shot learning, and semi-supervised learning.
Figure~\ref{fig:learning_strategy} illustrates the brief pipeline of different learning strategies.
In the following section, we review some representative methods for each strategy.

\vspace{1mm}
\noindent\textbf{Supervised Learning (SL)}
For shadow removal, supervised learning refers to the model training with paired shadow and shadow-free images captured in the same scene, but under different illumination conditions.
The pioneering deep learning-based work DeShadowNet~\cite{qu2017deshadownet} propose an automatic and end-to-end deep neural network to unify shadow detection, umbra/penumbra region classification, and shadow removal, which directly learns the mapping function between the shadow image and its shadow matte.
Subsequently, Wang \etal~\cite{wang2018stacked} developed a stacked conditional generative adversarial network (ST-CGAN) to simultaneously learn shadow detection and shadow removal.
The ST-CGAN organizes all tasks in a way that allows them to focus on one task at a time during different stages, while also benefiting from mutual improvements through the exchange of information in both forward and backward directions.
Hu \etal~\cite{hu2019direction} solve the shadow removal problem by a new perspective, which employs the spatial recurrent neural network with the direction-aware attention mechanism to exploit the context information.
After that, Zhang \etal~\cite{zhang2020ris} design a coarse-to-fine pipeline with four generators as well as three discriminators to explore the residual and illumination information separately.
A dual hierarchically aggregation network, DHAN~\cite{cun2020towards}, is proposed for shadow removal.
The DHAN contains a series of growth dilated convolutions as the backbone and hierarchically aggregates multi-context features for attention and prediction.
Lin \etal~\cite{lin2020bedsr} specifically design BEDSR-Net for document image shadow removal and take advantage of specific properties of document images, consisting of the global background color and attention map estimation. 
Recently, a two-stage context-aware network~\cite{chen2021canet} is proposed, which transfers contextual information from non-shadow to shadow regions at different scales with a contextual patch matching module.

In comparison to jointly learning shadow detection and shadow removal to explore both the semantic and fine-structural levels,
some recent methods enjoy better performance and more lightweight models by relying on an external pre-trained shadow detector or manually annotated mask information.
A group of methods explicitly combine the physical shadow illumination model into the deep network to increase interpretability.
For instance,
Le \etal~\cite{le2019shadow} propose to integrate the linear illumination transformation into the deep network to model the shadow effects.
This work uses a regression model to estimate the scaling factor and additive constant parameters for linear transformation. 
Following that, Fu \etal~\cite{fu2021auto} re-formulate shadow removal as an exposure fusion challenge. They reconstruct shadow-free results by intelligently blending over-exposed shadow images with shadow images using an adaptive per-pixel kernel weight map.
Zhu \etal~\cite{zhu2022efficient} design an iterative algorithm and unfold pipeline to  
ensure the identity mapping among non-shadow regions and adaptively perform the fine-grained spatial mapping between shadow regions.
Moreover, the shadow images can be divided into shadow and non-shadow regions based on the pre-defined shadow mask, which is very effective prior for shadow removal.
A Mask-ShadowNet~\cite{he2021mask} is proposed, which includes a masked adaptive instance normalization layer with embedded aligners and processes the shadow and non-shadow regions separately by different statistics.
To ensure the consistency between shadow removal and generation, Zhu \etal~\cite{zhu2022bijective} propose a bijective mapping network (BMNet) to recover the underlying
background contents during the forward procedure under the guidance of a color map and shadow mask. 
Guo~\etal~\cite{guo2023shadowformer} propose ShadowFormer to exploit the illumination information from non-shadow regions to help shadow regions restoration via the customized transformer blocks.

\vspace{1mm}
\noindent\textbf{Semi-Supervised Learning (SSL)}
In recent years, semi-supervised learning has emerged as an approach to leverage the strengths of both supervised learning and unsupervised learning.
It leverages both paired data and unpaired data to boost the model performance and improve generalization ability.
Ding~\etal~\cite{ding2019argan} propose a semi-supervised deep attentive recurrent generative adversarial network (ARGAN) to progressively detect and remove shadows.
The generator of ARGAN performs multiple progressive steps in a coarse-to-fine manner, serving the dual purpose of generating shadow attention maps and recovering the shadow-removal images.
The semi-supervised strategy is implemented by incorporating a significant number of unsupervised shadow images available online during the training process, which achieves better robustness and performance to complicated environment.

\vspace{1mm}
\noindent\textbf{Unsupervised Learning (UL)}
While supervised learning methods have shown remarkable success in various scenarios, they heavily rely on large amounts of paired training data, which can be costly and time-consuming to obtain.
Besides, training a deep model on paired data is typically designed to solve specific tasks, which struggle to adapt to new or out-of-distribution cases without retraining.
To address this issue, inspired by unsupervised image translation methods, Hu~\etal~\cite{hu2019mask} propose the mask guided cycle Generative Adversarial Network (GAN) using dual-generator, named Mask-ShadowGAN, to employ the unpaired data for shadow removal.
Similarly, Liu~\etal~\cite{liu2021shadow} also follow the cycle consistency pipeline while integrate the lightness information into the generator.
Different from the previous works requiring the pre-defined shadow mask, the DC-ShadowNet~\cite{jin2021dc} is proposed to employ the shadow/shadow-free domain classifier as the guidance and a series of loss functions to enable the generator focus on the shadow regions.
He~\etal~\cite{he21unsupervised} focus on portrait shadow removal instead of general one, which proposes to solve shadow removal task as a layer decomposition problem and leverages the generative facial priors.

\vspace{1mm}
\noindent\textbf{Zero-Shot Learning (ZSL)}
Supervised learning, and unsupervised learning methods exhibit limitations in terms of their ability to generalize or undergo stable training. To address these challenges, the concept of zero-shot learning has been introduced in low-level vision tasks. In this context, zero-shot learning refers to the approach of learning image enhancement solely from testing images, without the need for paired or unpaired training data. It is important to note that the definition of zero-shot learning in low-level vision tasks differs from its definition in high-level visual tasks. The goal is to emphasize that the method does not rely on specific training data arrangements, aiming to overcome the limitations and instability associated with other learning methods in low-level vision tasks~\cite{shocher2018zero,guo2020zero}.
Shadows, in fact, only form partial degradation in images, while their non-shadow regions provide rich structural and illumination information for shadow regions' restoration.
Based on that, Le~\etal~\cite{le2020shadow} propose to crop image patches with and without shadows from shadow images as the unpaired training samples, which can be trained using shadow images themselves.
Then they introduce a set of physics-based constraints that enables the adversarial training.
After that, Liu~\etal~\cite{liu2021from} propose the G2R-ShadowNet to leverage shadow generation to simulate shadow degradation in non-shadow regions, leading to paired data for training.

\begin{figure}
    \centering
    \includegraphics[width=7.5cm]{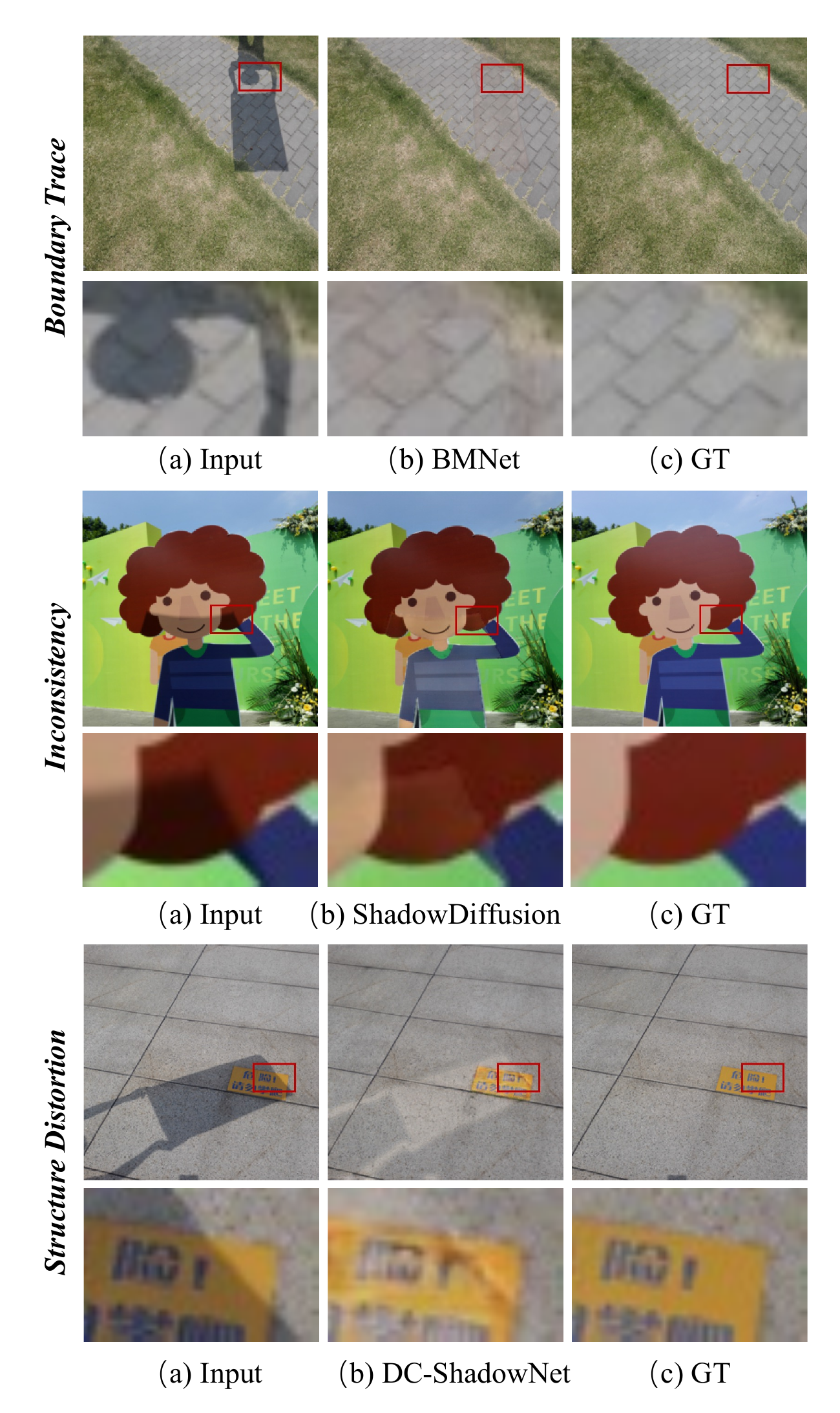}
    \caption{\bihan{Illustration of the three major artifacts/challenges in deep learning-based shadow removal, \ie, boundary trace (top), color and illumination inconsistency (middle), and structure distortion (bottom). Results by the representative method for each challenge are shown, \ie, BMNet~\cite{zhu2022bijective}, ShadowDiffusion~\cite{guo2022shadowdiffusion}, and DC-ShadowNet~\cite{jin2021dc}, respectively.}}
    \label{fig:challenge}
\end{figure}

\subsection{Challenges and Analysis}\label{sec:challenge}
The goal of single-image shadow removal is to eliminate shadows from images while maintaining or restoring the original appearance of the scene.
However, there are a few challenges to accomplishing the goal (refer to Figure~\ref{fig:challenge}):

\begin{figure*}[tbp]
    \centering
    \vspace{-3mm}
    \subfigure[Learning Strategy]{
    \begin{minipage}[b]{0.23\linewidth}
        \includegraphics[width=\linewidth, trim={1cm 0.2cm 3cm 0.2cm}, clip]{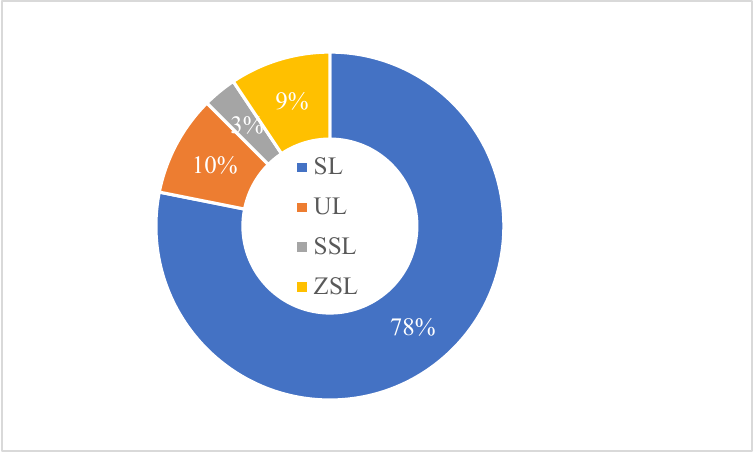}\vspace{-0.4cm}
    \end{minipage}
    }
    \hspace{-0.3cm}
    \subfigure[Network Structure]{
    \begin{minipage}[b]{0.23\linewidth}
         \includegraphics[width=\linewidth, trim={1cm 0.2cm 3cm 0.2cm}, clip]{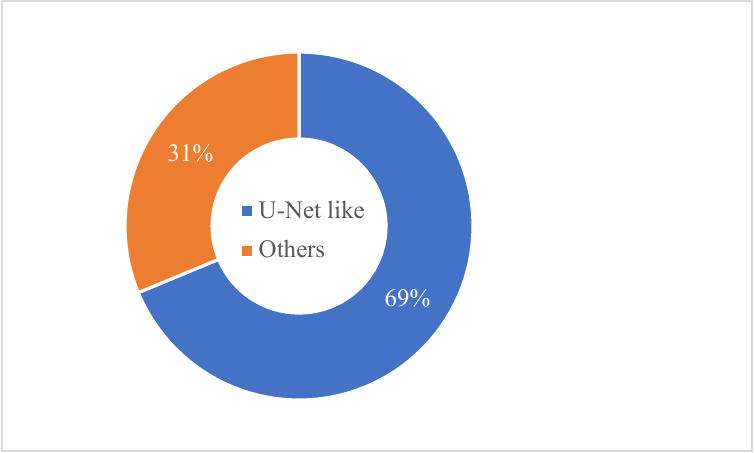}\vspace{-0.4cm}
    \end{minipage}
    }
    \hspace{-0.3cm}
    \subfigure[Mask Input]{
    \begin{minipage}[b]{0.23\linewidth}
         \includegraphics[width=\linewidth, trim={1cm 0.2cm 3cm 0.2cm}, clip]{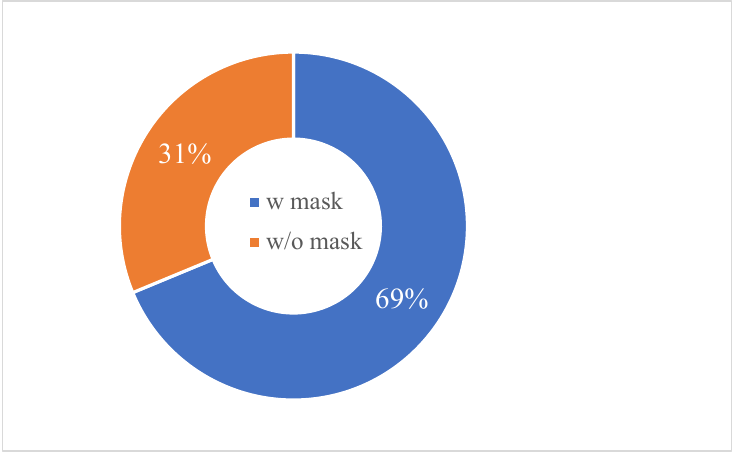}\vspace{-0.4cm}
    \end{minipage}
    }
    \hspace{-0.3cm}
    \subfigure[Physic Model]{
    \begin{minipage}[b]{0.23\linewidth}
           \includegraphics[width=\linewidth, trim={1cm 0.2cm 3cm 0.2cm}, clip]{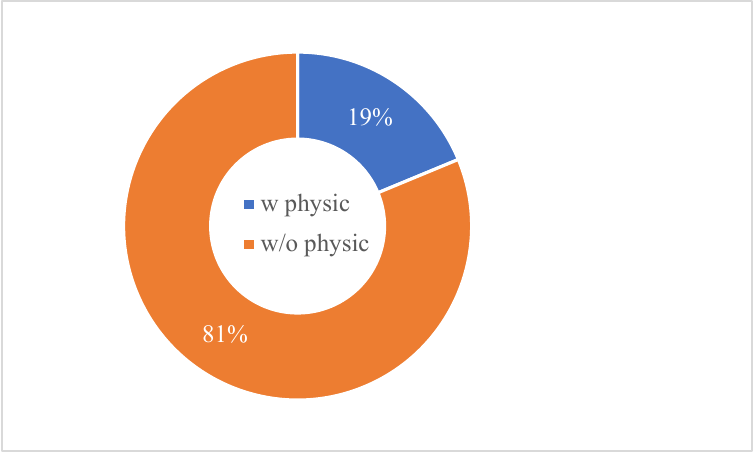}\vspace{-0.4cm}
    \end{minipage}
    }
        \subfigure[Training Dataset]{
    \begin{minipage}[b]{0.23\linewidth}
        \includegraphics[width=.9\linewidth, trim={0.07cm 2.2cm 0.1cm 0.1cm}, clip]{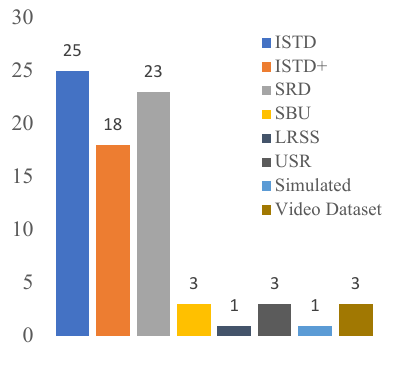}\vspace{-0.4cm}
    \end{minipage}
    }
    \hspace{-0.3cm}
    \subfigure[Testing Dataset]{
    \begin{minipage}[b]{0.23\linewidth}
      \includegraphics[width=.95\linewidth, trim={0.1cm 2.8cm 0.1cm 0.1cm}, clip]{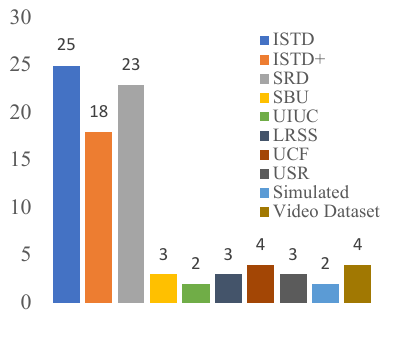}\vspace{-0.4cm}
    \end{minipage}
    }
    \hspace{-0.3cm}
    \subfigure[Loss Function]{
    \begin{minipage}[b]{0.23\linewidth}
        \includegraphics[width=.95\linewidth, trim={0.1cm 2.2cm 0.1cm 0.1cm}, clip]{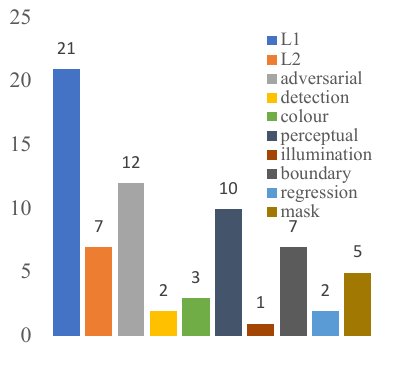}\vspace{-0.4cm}
    \end{minipage}
    }
    \hspace{-0.3cm}
    \subfigure[Evaluation Metric]{
    \begin{minipage}[b]{0.23\linewidth}
         \includegraphics[width=.95\linewidth, trim={0.1cm 1.6cm 0.1cm 0.1cm}, clip]{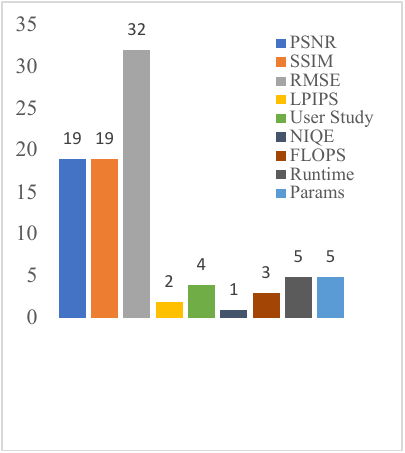}\vspace{-0.4cm}
    \end{minipage}
    }
    \caption{\bihan{A statistical analysis of the number of deep learning-based shadow removal methods, according to their} learning strategy, network characteristic, mask input, physic model, training dataset, testing dataset, loss function, and evaluation metric.}    
    \label{fig:statistic}
\end{figure*}






\begin{itemize}
    \item \textbf{Trace-less result.} 
    The patterns of shadows are characterized by their arbitrary nature, diversity, and occasionally, by their highly intricate trace structures, which present substantial challenges in achieving ``trace-less'' image restoration. 
    Consequently, the fabrication of a shadow mask that accurately and meticulously captures the shadow patterns proves to be a formidable task. 
    Moreover, the variations in illumination around the shadow boundaries are pronounced, complicating the modeling of the sophisticated and abrupt changes in these boundary regions. 
    As a consequence of these complexities, the removal of shadows, particularly those with hard, sharp edges, is prone to the generation of artifacts along the edges where the shadow interfaces with the non-shadowed areas.
   In addressing this challenge, conventional approaches~\cite{fredembach2004fast,gryka2015learning} typically involve smoothing the gradient around the boundary. Similarly, deep learning-based methods~\cite{jin2021dc,guo2023boundary} have employed customized Total Variation (TV) loss functions in boundary regions to mitigate boundary artifacts.

    \item \textbf{Illumination and color consistency.} 
    Shadows represent more than just areas of darker pixels; they encompass complex visual phenomena characterized by variations in color, texture, and brightness. These characteristics are influenced by the properties of both the light source and the surfaces they interact with. Furthermore, shadows, despite their impact on image quality, constitute only partial degradation within images. Conversely, the non-shadowed regions offer rich and accurate color and illumination information, serving as a reference standard.
    As a result, the goal of shadow removal is not only to enhance the shadow regions but also to pursue consistent illumination and color compared to non-corrupted (non-shadow) regions.
    Several existing deep learning-based methodologies have endeavored to expand the receptive field to incorporate valuable information from non-shadowed regions as guidance. This has been achieved through various techniques, including the fusion of multi-level features~\cite{qu2017deshadownet}, employment of dilated convolutions~\cite{cun2020towards}, utilization of patch matching algorithms~\cite{chen2021canet}, and incorporation of transformer mechanisms~\cite{guo2023shadowformer}.
    

    \item \textbf{Structure distortion.} 
    Data fidelity is a prevalent concern in various image restoration tasks, including shadow removal, where the aim is to produce high-fidelity results that closely resemble the original scene.
Particularly within the context of shadow removal, unique challenges arise.    On the one hand, objects with intricate textures or reflective surfaces can exhibit complex shadow patterns that are challenging to remove without affecting the underlying structure.
    The removal process must distinguish between shadow-related texture and genuine object details to avoid distortion.
    Besides, the boundaries between shadowed and non-shadowed areas can be ambiguous or ill-defined, particularly in scenes with diffuse lighting or complex geometry. 
    Errors in delineating these boundaries can lead to inaccuracies in shadow removal and subsequent structure distortion.
    On the other hand, some works~\cite{wang2018stacked,hu2019mask} applied adversarial losses for training shadow removal networks and tried to model the image distribution explicitly. However, these models usually hallucinate unrealistic image contents and structures.
To mitigate this issue, some existing methods have adopted perceptual loss techniques to ensure structural consistency between input shadow images and the resulting outputs.    
\end{itemize}

\section{Technical Review and Discussion}\label{sec:technical}

In this section, we review and discuss the technical details and employed priors designed to address the aforementioned challenges in Section~\ref{sec:challenge} from different perspectives, such as network structure, integrating physical illumination models, exploiting shadow detection for removal, and using deep generative models for shadow removal. We then summarize the widely used loss functions, training and testing datasets, and evaluation metrics.

\subsection{Network Structure}

\begin{figure*}
\vspace{-3mm}
    \centering
    \includegraphics[width=16cm]{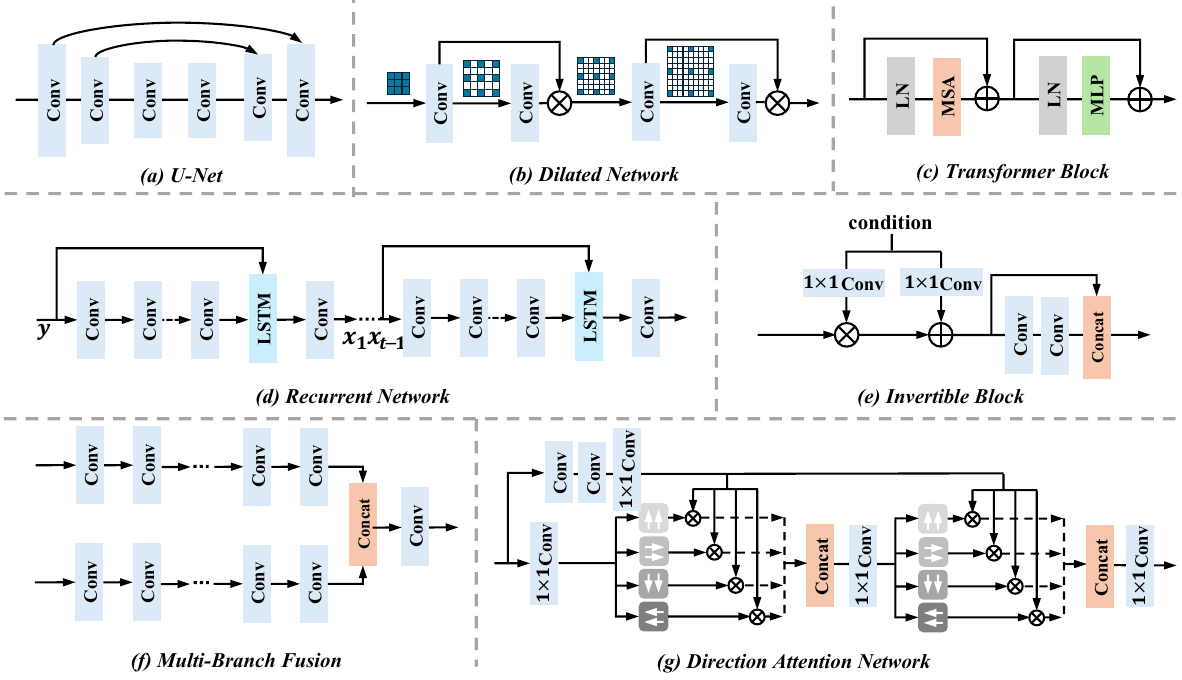}
    \caption{\bihan{The network architectures of the building blocks that are widely used in deep shadow removal algorithms.}}
    \label{fig:blocks}
\end{figure*}

As deep learning techniques advance, there's a noticeable trend in newly proposed methods to incorporate more sophisticated basic blocks with enhanced modeling capabilities. These blocks are subsequently integrated to build more complex shadow removal networks. In the following section, we provide detailed designs and rationales for various basic blocks employed in deep learning-based shadow removal networks, as illustrated in Figure~\ref{fig:blocks}.

\vspace{1mm}
\noindent\textbf{U-Net and Multi-Scale}
Addressing shadow removal as a partial corruption problem brings forth its own set of unique challenges, prompting us to emphasize the importance of maximizing the receptive field of the model to encompass as much non-corrupted context information as possible. 
The widely adopted U-Net-like structure \cite{ronneberger2015u} (Figure~\ref{fig:blocks}(a)) serves as a fundamental backbone in shadow removal tasks, representing approximately 69\% of the approaches, as depicted in Figure~\ref{fig:statistic}(b).
This architecture has garnered widespread usage due to its ability to preserve both low-level structural details and high-level semantic information, facilitating high-quality restoration of shadow-free images while accurately identifying shadow regions. 
Moreover, alternative structural designs incorporate multi-scale information using various dilated kernels (Figure~\ref{fig:blocks}(b)) or by leveraging different feature maps extracted from pre-trained models, such as VGG16 (Figure~\ref{fig:blocks}(f)). 
This approach also enables the capture of diverse information across multiple levels and effectively enlarging the receptive field. 


\vspace{1mm}
\noindent\textbf{One-Stage \textit{v.s.} Multi-Stage}
A few methods incorporate the use of two-stage or multi-stage pipelines for shadow removal.
Such a strategy enables the application of fine-grained constraints at different stages of the pipeline, facilitating more precise and effective restoration.
By dividing the whole restoration process into multiple stages, each stage can conduct its specific set of constraints.
For instance, certain techniques~\cite{le2019shadow,fu2021auto} integrate an additional refinement module following the initial processing stage. This refinement step aims to further enhance the structural details or mitigate any artifacts present in the output produced by the preceding stage.
Nevertheless, in contrast to single-stage methods, they often demand double or even multiple times the number of parameters and inference time.

\vspace{1mm}
\noindent\textbf{CNN \textit{v.s.} Transformer}
The era of deep learning-based shadow removal began in the year 2017 with the work of Qu~\etal~\cite{qu2017deshadownet}.
They introduced a pioneering multi-branch fusion architecture (Figure~\ref{fig:blocks}(f)) designed to independently extract both local and global information using cascaded CNNs across different layers.
Following this, a series of CNN-based networks~\cite{le2019shadow,le2020shadow,zhu2022efficient,cun2020towards,jin2021dc} are constructed to improve the performance of shadow removal.
After that, the transformer mechanism~\cite{vaswani2017attention} capitalizes on long-range dependencies within contexts, leading to notable enhancements across various vision tasks, including image restoration and enhancement. In scenarios involving partial corruption, such as shadow removal, transformer-based networks exhibit greater potential due to the limited availability of corrupted contextual information to guide restoration. As a result, transformer-based models (Figure~\ref{fig:blocks}(c)) are becoming increasingly prevalent in the field of shadow removal~\cite{yu2023cnsnet,guo2023shadowformer}.


\vspace{1mm}
\noindent\textbf{Recurrent and Recursive} 
Recurrent neural networks (RNNs) can be used in an iterative manner to refine the restoration results over multiple iterations. 
This iterative refinement process allows RNN-based models to effectively address complex restoration tasks that require fine-grained adjustments and corrections.
Certain existing methods~\cite{ding2019argan} have employed recurrent networks, depicted in Figure~\ref{fig:blocks}(d), to gradually detect and eliminate shadows using a sequential process. Hence, employing a recurrent unit like Long Short-Term Memory (LSTM) can ensure the preservation of valuable and detailed information, thereby enhancing the accuracy of detected shadow regions and producing increasingly realistic shadow-removal images.
By dividing the process into multiple stages, each with its specific set of constraints, the system can better address various aspects of shadow removal, resulting in enhanced performance and accuracy.

\vspace{1mm}
\noindent\textbf{Attention Mechanism}
Shadows often occur in specific regions of an image, and the attention mechanism allows the model to selectively focus on these shadowed area.
By attending to relevant parts of the image, the model can effectively target shadow regions for removal while preserving non-shadowed areas.
Some methods~\cite{wang2018stacked,hu2019mask} directly serve the shadow mask as the attention for the removal.
Others predict a learnable attentive map as the auxiliary module by considering the direction (Figure~\ref{fig:blocks}(g)) or recurrent attention~\cite{zhu2018bidirectional}.


\begin{table*}
		\rowcolors{1}{gray!20}{white}
		\centering
  \caption{
Summary of key characteristics of notable deep learning-based methods. ``Physic'' denotes whether the methods integrate physical models.  ``Mask'' shows whether the methods require a mask as input for inference. }
		\vspace{-6pt}
		\label{table:methods}
		\begin{threeparttable}
			\resizebox{1\textwidth}{!}{
				\setlength\tabcolsep{2pt}
				\renewcommand\arraystretch{1}
				\begin{tabular}{c|r||c|c|c|c|c|c|c|c|c}
					\hline
					&\textbf{Method}~~~~~~~~~&\textbf{Learning} &\textbf{Network Structure} &\textbf{Loss Function}
					&\textbf{Training Data} & \textbf{Testing Data} &\textbf{Evaluation Metric}  &\textbf{Platform} &\textbf{Physic}&\textbf{Mask}\\
					\hline
					\hline
					\multirow{1}{*}{\rotatebox{90}{\textbf{2017}}}
					&DeShadowNet~\cite{qu2017deshadownet} &SL &\tabincell{c}{multi-branch fusion \\semantic network, end-to-end} &MSE loss (log space) & SRD &  UIUC LRSS SRD & RMSE SSIM (LAB space) & Caffe &&\\
					\hline
					\hline
					\multirow{1}{*}{\rotatebox{90}{\textbf{2018}}}
					& ST-CGAN~\cite{wang2018stacked} & SL &   \tabincell{c}{U-Net like network\\ joint learning \\GAN, end-to-end}  & \tabincell{c}{adversarial loss \\ reconstruction loss \\ detection loss}& ISTD SBU &  SBU UCF ISTD & RMSE (LAB space) & PyTorch & &\\
				
					\hline
					\hline
					\multirow{1}{*}{\rotatebox{90}{\textbf{2019}}}
					& SP+M-Net~\cite{le2019shadow} & SL &   \tabincell{c}{U-Net like network\\ regression model(ResNeXt) \\two subnetworks}  & \tabincell{c}{regression loss \\ reconstruction loss}& ISTD SBU &  SBU UCF ISTD & RMSE (LAB space) & PyTorch & \checkmark & \checkmark\\
             & DSC~\cite{hu2019direction} & SL &   \tabincell{c}{spatial RNN\\ direction-aware attention}  & \tabincell{c}{colour loss \\ reconstruction loss}& ISTD SBU &  SBU UCF ISTD & RMSE (LAB space) & Caffe & & \\
					& Mask-ShadowGAN~\cite{hu2019mask} & UL & \tabincell{c}{U-Net like network\\ GAN \\ dual-generator} & \tabincell{c}{cycle-consistency loss \\ adversarial loss \\ identity loss} & SRD ISTD USR & SRD ISTD USR & \tabincell{c}{RMSE (LAB space)\\User Study} & PyTorch & &\checkmark\\
									& ARGAN~\cite{ding2019argan} & SSL & \tabincell{c}{ GAN \\ recursive network} & \tabincell{c}{detection loss \\ adversarial loss \\ reconstruction loss \\ perceptual loss} & SRD ISTD & SRD ISTD & \tabincell{c}{RMSE (LAB space)} & TensorFlow & &\\
					\hline
					\hline
					\multirow{1}{*}{\rotatebox{90}{\textbf{2020}}}
					&RIS-GAN~\cite{zhang2020ris} &SL &\tabincell{c}{U-Net like network \\four subnetworks \\ GAN joint discriminator} &\tabincell{c}{perceptual loss \\ reconstruction loss ($\ell_1$) \\ residual loss illumination loss \\cross loss adversarial loss}
					&ISTD SRD &ISTD SRD &\tabincell{c}{RMSE (LAB space)\\User Study} &TensorFlow & &\\
					&DHAN~\cite{cun2020towards}  &SL &\tabincell{c}{dilated convolution, GAN \\ hierarchical attention \\spatial pooling pyramid } &\tabincell{c}{ perceptual loss\\binary cross entropy loss \\ adversarial loss}
					&\tabincell{c}{ISTD SRD \\ simulated} &\tabincell{c}{ISTD SRD \\ simulated} & \tabincell{c}{LPIPS SSIM\\ PSNR RMSE} &TensorFlow & &\\
					&Le \etal~\cite{le2020shadow}& ZSL &\tabincell{c}{U-Net like network \\ regression network \\ GAN} &\tabincell{c}{matting loss smoothness loss\\boundary loss\\adversarial loss}
					&\tabincell{c}{ISTD \\ Video Dataset} &\tabincell{c}{ISTD \\ Video Dataset} &RMSE (LAB space) &\tabincell{c}{TensorFlow}&\checkmark & \checkmark\\
					\hline
					\hline
					\multirow{1}{*}{\rotatebox{90}{\textbf{2021}}}
					&Fu \etal~\cite{fu2021auto}  &SL &\tabincell{c}{regression model \\ multi-stage, fusion network \\ U-Net like network} &\tabincell{c}{reconstruction loss ($\ell_1$)\\boundary gradient loss}
					&\tabincell{c}{SRD ISTD ISTD+} &\tabincell{c}{SRD ISTD ISTD+} &\tabincell{c}{RMSE (LAB space)}&PyTorch &\checkmark & \checkmark\\
					&Mask-ShadowNet~\cite{he2021mask}& SL &\tabincell{c}{U-Net like network \\ adaptive instance normalization \\
					end-to-end} &\tabincell{c}{perceptual loss}
					&\tabincell{c}{ISTD} &ISTD &RMSE (LAB space) &\tabincell{c}{PyTorch}& & \checkmark\\
					&CANet~\cite{chen2021canet} & SL &\tabincell{c}{DenseNet \\ patch matching mechanism \\ contextual feature transfer\\ DenseUNet \\ multi-stage} &\tabincell{c}{regression loss \\ reconstruction loss ($L_2$)\\ cross-entropy loss\\perceptual loss\\gradient loss}
					&SRD ISTD &SRD ISTD &RMSE (LAB space)  &PyTorch & &\\
					&G2R-ShadowNet~\cite{liu2021from}  & ZSL & \tabincell{c}{U-Net like network \\GAN \\three subnetworks} &\tabincell{c}{adversarial loss \\ identity loss \\cycle-consistency loss \\ refinement loss area loss}
					&\tabincell{c}{ISTD \\ Video Dataset} &\tabincell{c}{ISTD \\ Video Dataset} &\tabincell{c}{PSNR SSIM\\RMSE (LAB space) } &PyTorch & & \checkmark\\
					&LG-ShadowNet~\cite{liu2021shadow}  & UL &\tabincell{c}{multi-stage \\ GAN \\ U-Net like network} &\tabincell{c}{identity loss\\cycle-consistency loss \\ adversarial loss \\ colour loss}
					&ISTD ISTD+ USR &ISTD ISTD+ USR &\tabincell{c}{PSNR SSIM User Study\\RMSE (LAB space) \\ SSEQ NIQE DBCNN  \\Runtime \#P FLOPs} &PyTorch && \checkmark\\
					&DC-ShadowNet~\cite{jin2021dc}  & UL &\tabincell{c}{U-Net like network \\GAN} &\tabincell{c}{chromaticity loss\\perceptual loss \\ classification loss \\boundary smoothness loss \\ adversarial loss identity loss \\ reconstruction consistency loss}
					&\tabincell{c}{SRD ISTD+ ISTD \\ USR LRSS} & \tabincell{c}{SRD ISTD+ ISTD \\ USR LRSS} &\tabincell{c}{PSNR RMSE (LAB space)} &PyTorch &&\\
					\hline
					\hline
					\multirow{1}{*}{\rotatebox{90}{\textbf{2022}}}
					&Zhu \etal~\cite{zhu2022efficient}  &SL &\tabincell{c}{U-Net like network \\unfolding \\recursive network} &\tabincell{c}{reconstruction loss\\regularization loss}
					&\tabincell{c}{ISTD SRD} &\tabincell{c}{ISTD SRD} &\tabincell{c}{PSNR SSIM\\ RMSE (LAB space) \\ Runtime \#P FLOPs}&PyTorch &\checkmark & \checkmark\\
					&BMNet~\cite{zhu2022bijective}&SL &invertible network &\tabincell{c}{colour loss \\ reconstruction loss ($\ell_1$)}
					&\tabincell{c}{ISTD ISTD+ SRD} &\tabincell{c}{ISTD ISTD+ SRD} &\tabincell{c}{ PSNR SSIM\\RMSE (LAB space)} &\tabincell{c}{PyTorch}&& \checkmark\\
					&CNSNet~\cite{yu2023cnsnet} & SL &\tabincell{c}{U-Net like network \\ transformer \\ adaptive normalization} &\tabincell{c}{reconstruction loss ($\ell_1$) \\ perceptual loss \\ boundary gradient loss \\ soft mask loss}
					&\tabincell{c}{ISTD ISTD+ SRD} &\tabincell{c}{ISTD ISTD+ SRD} &\tabincell{c}{ PSNR SSIM\\RMSE (LAB space)} &PyTorch && \checkmark\\
					& SG-ShadowNet~\cite{wan2022style}  & SL & \tabincell{c}{U-Net like network \\style transfer \\ prototypical normalization} &\tabincell{c}{reconstruction loss ($\ell_1$) \\ area loss \\spatial consistency loss}
					&\tabincell{c}{ISTD+ SRD} &\tabincell{c}{ISTD+ SRD \\ Video Dataset} &\tabincell{c}{LPIPS RMSE (LAB space)} &PyTorch & & \checkmark\\
										\hline
					\hline
					\multirow{1}{*}{\rotatebox{90}{\textbf{2023}}}
	
					&BA-ShadowNet~\cite{niu2022boundary}&SL &\tabincell{c}{encoder-decoder \\ multi-branch} &\tabincell{c}{boundary loss\\ similarity loss\\ shadow-free consistency loss}
					&\tabincell{c}{ISTD+ SRD} &\tabincell{c}{ISTD+ SRD} &\tabincell{c}{PSNR SSIM\\RMSE (LAB space)} &\tabincell{c}{PyTorch}&&\checkmark\\
					&DMTN~\cite{liu2023decoupled} & SL &\tabincell{c}{encoder-decoder \\ multi-branch} &\tabincell{c}{perceptual loss\\reconstruction loss ($\ell_1$)\\adversarial loss}
					&\tabincell{c}{ISTD ISTD+ SRD} &\tabincell{c}{ISTD ISTD+ SRD} &\tabincell{c}{PSNR SSIM\\RMSE (LAB space)} &PyTorch & \checkmark &\\
					&ShadowFormer~\cite{guo2023shadowformer}  &SL &\tabincell{c}{ transformer\\ encoder-decoder} &\tabincell{c}{reconstruction loss ($\ell_1$)}
					&\tabincell{c}{ISTD ISTD+ SRD} &\tabincell{c}{ISTD ISTD+ SRD} &\tabincell{c}{PSNR SSIM\\  \#P RMSE (LAB space)}&PyTorch & &\checkmark\\
					&SHARDS~\cite{sen2023shards}  &SL &\tabincell{c}{U-Net like network \\two-stage } &\tabincell{c}{adversarial loss\\perceptual loss}
					&ISTD SFHQ &ISTD SFHQ &RMSE (LAB space)  &PyTorch && \checkmark\\
					&LRA\&LDRA~\cite{yucel2023lra}  &SL &\tabincell{c}{U-Net like network \\residual learning } &\tabincell{c}{reconstruction loss ($\ell_1$)}
					&ISTD+ SRD &ISTD+ SRD &\tabincell{c}{PSNR SSIM\\ RMSE (LAB space) \\ Runtime \#P FLOPs}  &PyTorch & &\checkmark\\
				&ShadowDiffusion~\cite{guo2022shadowdiffusion}  &SL &\tabincell{c}{diffusion model \\ unfolding} &\tabincell{c}{reconstruction loss\\ mask loss}
					&\tabincell{c}{ISTD ISTD+ SRD} &\tabincell{c}{ISTD ISTD+ SRD} &\tabincell{c}{PSNR SSIM\\  RMSE (LAB space)}  &PyTorch & \checkmark &\checkmark\\
				&FSR-Net~\cite{yu2023fsr}  &SL &\tabincell{c}{multi-stage \\ U-Net like network} &\tabincell{c}{reconstruction loss\\ mask loss}
					&\tabincell{c}{ISTD ISTD+ SRD} &\tabincell{c}{ISTD ISTD+ SRD} &\tabincell{c}{PSNR SSIM\\  RMSE (LAB space)}  &PyTorch & &\checkmark \\
				&Inpaint4Shadow~\cite{li2023leveraging}  &SL &\tabincell{c}{fusion network \\ Resnet, inpainting network} &\tabincell{c}{reconstruction loss\\ mask loss}
					&\tabincell{c}{ISTD ISTD+ SRD} &\tabincell{c}{ISTD ISTD+ SRD} &\tabincell{c}{PSNR SSIM\\  RMSE (LAB space)}  &PyTorch &  &\checkmark\\
				&BCDiff~\cite{guo2023boundary}  &ZSL &\tabincell{c}{diffusion model \\ decomposition model} &\tabincell{c}{reconstruction loss\\ boundary loss}
					&\tabincell{c}{ISTD+ Video} &\tabincell{c}{ISTD+ Video} &\tabincell{c}{PSNR SSIM\\  RMSE (LAB space)}  &PyTorch &  &\checkmark\\
     	\hline
					\hline
					\multirow{1}{*}{\rotatebox{90}{\textbf{2024}}}
	
					&Liu~\etal~\cite{liu2024recasting}&SL &\tabincell{c}{encoder-decoder \\ multi-branch} &\tabincell{c}{boundary loss\\ similarity loss\\ shadow-free consistency loss}
					&\tabincell{c}{ISTD+ SRD} &\tabincell{c}{ISTD+ SRD} &\tabincell{c}{PSNR SSIM\\RMSE (LAB space)} &\tabincell{c}{PyTorch}&&\checkmark\\
					&DeS3~\cite{jin2022des3} & SL &\tabincell{c}{diffusion model \\ attention \\classifier guidance} &\tabincell{c}{perceptual loss\\reconstruction loss ($\ell_1$)\\adversarial loss}
					&\tabincell{c}{ISTD ISTD+ SRD} &\tabincell{c}{ISTD ISTD+ SRD \\ LRSS UCF UIUC} &\tabincell{c}{PSNR SSIM User Study\\RMSE (LAB space)} &PyTorch &  &\\
					&Mei~\etal~\cite{mei2024latent}  &SL &\tabincell{c}{ latent diffusion\\ multi-stage} &\tabincell{c}{reconstruction loss}
					&\tabincell{c}{ISTD ISTD+ SRD} &\tabincell{c}{ISTD ISTD+ \\SRD DeSOBA} &\tabincell{c}{PSNR SSIM Runtime\\  \#P RMSE (LAB space)}&PyTorch & &\checkmark\\
     &HomoFormer~\cite{Xiao_2024_CVPR}  &SL &\tabincell{c}{ transformer\\ U-Net like network} &\tabincell{c}{reconstruction loss}
					&\tabincell{c}{ISTD+ SRD} &\tabincell{c}{ISTD+ SRD} &\tabincell{c}{PSNR SSIM Runtime\\  \#P RMSE (LAB space)}&PyTorch & &\checkmark\\
					\hline
				\end{tabular}
			}
		\end{threeparttable}
	\end{table*}

\subsection{Integrating Physical Illumination Model}

The physical properties of shadows are effective prior to shadow removal, which guarantees that the networks would learn physically plausible transformations.

\vspace{1mm}
\noindent\textbf{Parameter estimation.}
Some studies~\cite{le2019shadow,le2020shadow} employ a linear function to model the relationship between illuminated and shadowed pixels. They assume this linear relationship remains consistent during the camera's color acquisition process~\cite{finlayson2016rank}. Consequently, the illuminated image can be represented as a linear function of its shadowed values:
\begin{align}
& \mathbf{I}_{l i t}=\omega \circ \mathbf{I}_s+b, \\
& 
\chong{\hat{\mathbf{I}}_{sf}}=\boldsymbol{\alpha} \circ \mathbf{I}_s+(\mathbf{1}-\boldsymbol{\alpha}) \circ \mathbf{I}_{l i t},
\end{align}
where $\omega=\left[\omega_R, \omega_G, \omega_B\right], b=\left[b_R, b_G, b_B\right]$ for different color channels.
 $b$ represents the camera's response to direct illumination, while $\omega$ denotes the attenuation factor of ambient illumination for a given pixel in a particular color channel.
 With this assumption in mind, they employ a deep regression model to estimate the parameters $\omega$ and $b$ based on input data comprising both shadow and shadow-free images.
 Subsequently, an additional network, specialized for matte prediction, is applied to generate the pixel-wise shadow matte $\boldsymbol{\alpha}$. 
 This matte effectively combines information from the illuminated and shadowed regions of the images. 

\vspace{1mm}
\noindent\textbf{Deep unrolling.}
Another group of works~\cite{zhu2022efficient,guo2022shadowdiffusion} employs deep unrolling to enhance the interpretability of models. This popular and powerful approach incorporates known physical priors into deep networks, achieving superior performance across various image restoration problems~\cite{zhang2020deep, wang2024progressive, zheng2022unfolded, wang2023deep, kong2021deep, mou2022deep}.
The shadow formation model~(\ref{eq:shadow_formation}) can be re-formulated as
\begin{equation}
    \mathbf{I}_{s} = \mathbf{A}\circ \mathbf{I}_{sf}\;.
\end{equation}
Based on that, the shadow removal can be formulated as a degradation prior guided model, where regularization terms $\mathcal{R}(\cdot)$ are inferred
by the learnable conditional deep CNN networks~\cite{zhu2022efficient} or generative diffusion model~\cite{guo2022shadowdiffusion} as follows:
\begin{equation}
   \hat{\mathbf{I}}_{sf} =  \argmin_{\mathbf{I}_{sf}} \|\mathbf{I}_{s} - \mathbf{A}\circ\mathbf{I}_{sf}\|_F^2 + \mathcal{R}(\mathbf{I}_{sf})\;.
\end{equation}



\subsection{Exploiting Shadow Detection for Removal}
\begin{figure}
    \centering
    \includegraphics[width=7.5cm]{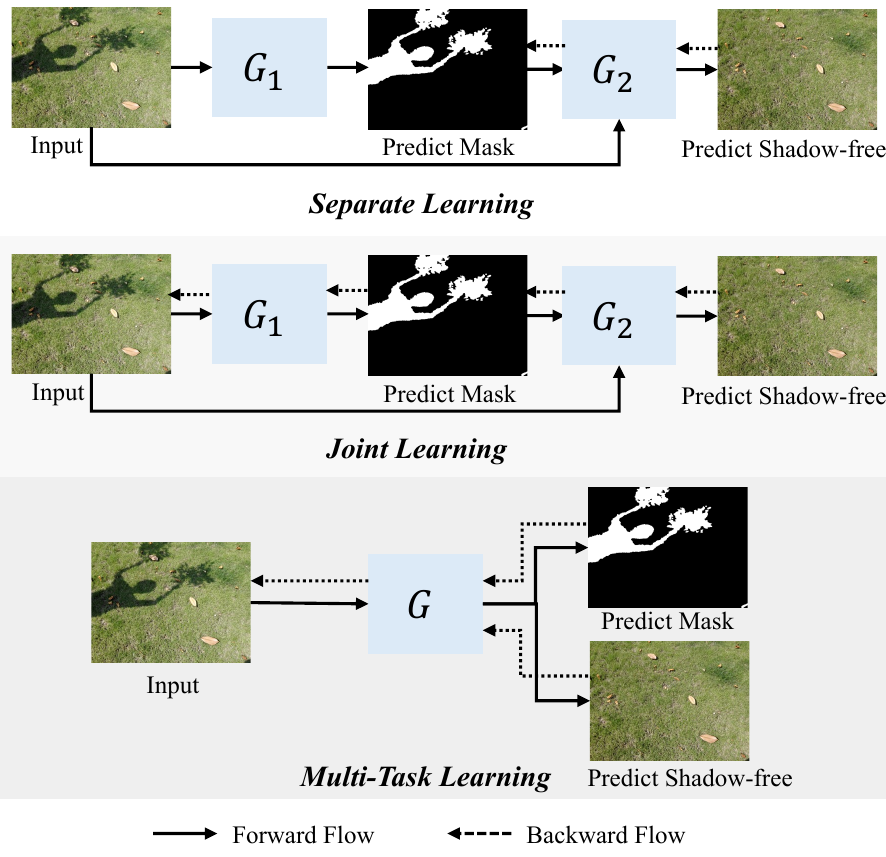}
    \caption{\bihan{Illustration of the three strategies of combining shadow removal and detection,} \ie, separate learning, joint learning, and multi-task learning strategies.}
    \label{fig:detection}
\end{figure}

Shadow removal often relies on accurate shadow detection. Before removing shadows from an image, it's essential to identify where the shadows are present. Shadow detection algorithms help locate regions of shadow within an image, providing the necessary information for subsequent removal processes.
Most existing shadow removal methods leverage rich semantic and structural information learnt from shadow detection as prior knowledge, which can roughly be divided into three widely used frameworks as shown in Figure~\ref{fig:detection}:
\begin{enumerate}
    \item \textbf{Separate learning}: 
Most shadow removal techniques (refer to the methods that require mask input in Table~\ref{table:methods}) incorporate a pre-trained shadow detection model before the removal stage. This model generates a corresponding shadow mask for the input shadow image, which serves as a conditional input for the shadow removal network. The shadow detection model remains fixed and operates as a pre-processing step for the shadow removal process.

\item \textbf{Joint learning}: 
The adopted frozen shadow detection model may encounter a domain gap when tested on cases with distinct distributions from the training data. Inaccurate detection of shadows could misguide the subsequent shadow removal network. To address this, some approaches~\cite{wang2018stacked} have employed a joint training strategy, optimizing both the shadow detection and removal networks simultaneously. This allows for fine-tuning the shadow detector to the current data domain, effectively mitigating the domain gap issue.

\item \textbf{Multi-task learning}:  
Another category of shadow removal methods~\cite{cun2020towards} has adopted the multi-task learning framework to exploit shared representations and simultaneously output shadow mask and shadow-free results.
This approach minimizes redundancy in feature representation learning and parameter sharing, resulting in more efficient utilization of computational resources and decreased inference time.
    
\end{enumerate}


\subsection{Deep Generative Model for Shadow Removal}
Generative models, by design, learn the underlying data distribution of training images. 
This allows them to generate or reconstruct images that are close to the real shadow-free data distribution, effectively filling in missing textures or colors and avoiding amplified artifacts, such as over-smoothness. 
This is particularly useful in shadow removal, where the model must
understand the natural image prior to effectively removing shadows without leaving traces. 
Here, we primarily discuss two types of generative models that are widely used in shadow removal: Generative Adversarial Networks (GANs)~\cite{goodfellow2014generative} and diffusion models~\cite{ho2020denoising}.

\vspace{1mm}
\noindent\textbf{GAN-based methods.}
GAN-based models for shadow removal leverage the adversarial training framework to distinguish between shadow and non-shadow regions, aiming to generate shadow-free images that are indistinguishable from real, shadowless photos.
While the integration of adversarial loss in shadow removal models can mitigate some issues such as boundary artifacts, such as those proposed by Wang~\etal~\cite{wang2018stacked} and Hu~\etal~\cite{hu2019mask} necessitate careful adjustment during training.
Furthermore, these methods may exhibit tendencies to overfit specific visual features or data distributions, potentially leading to the generation of hallucinated content and artifacts.



\vspace{1mm}
\noindent\textbf{Diffusion-based methods.}
Recently, numerous diffusion models, including the widely recognized diffusion denoising probability model (DDPM)\cite{ho2020denoising}, have garnered significant interest in the field of low-level vision\cite{sohl2015deep,ho2020denoising,kawar2022denoising,saharia2022image,wang2023exposurediffusion,wang2024exploiting}.
Guo~\etal~\cite{guo2022shadowdiffusion} pioneered the integration of diffusion models into the domain of shadow removal. 
Subsequently, a series of methods~\cite{jin2022des3,guo2023boundary,mei2024latent} tend to employ the diffusion model as the backbone.
While diffusion-based methods have demonstrated their efficacy in producing realistic shadow-free results, they are often accompanied by a significant drawback: the time-consuming nature of the inference process, primarily attributable to the iterative sampling procedures involved.
In some applications, such as real-time processing, there may be strict requirements on the latency of inference.

\subsection{Loss Function}

The commonly adopted loss functions in shadow removal methods are the full-reference loss, \eg, reconstruction loss ($\chong{\ell}_1$, $\chong{\ell}_2$), and perceptual loss.
Additionally, depending on specific settings and formulations, some non-reference loss functions like boundary loss, adversarial loss, and color loss are also utilized.
We provide a detailed explanation of some representative loss functions as follows.

\vspace{1mm}
\noindent\textbf{Reconstruction Loss.}  
Reconstruction loss refers to the measure of dissimilarity between the ground truth image and the restored image produced by the model, which is typically computed by comparing the pixel values of the restored image with the corresponding pixel values of the clean image. Different reconstruction losses have their advantages and disadvantages~\cite{zhao2016loss,li2021low}. $\chong{\ell}_2$ loss, also known as mean squared error (MSE) loss, penalizes larger errors more than smaller errors. While $\chong{\ell}_2$ loss is effective at capturing global trends and minimizing overall reconstruction errors, it can sometimes blur or smooth out fine details due to its emphasis on larger errors.
On the other hand, $\chong{\ell}_1$ loss, also known as mean absolute error (MAE) loss, treats all errors equally regardless of their magnitude. 
It is less sensitive to outliers and can preserve colors and illuminance well, as it does not disproportionately penalize local deviations in the image structure. Consequently, $\chong{\ell}_1$ loss can result in sharper edges and better preservation of high-frequency details. Structural Similarity Index (SSIM) loss is designed to preserve the structure and texture of the image, leading to visually pleasing results.

\vspace{1mm}
\noindent\textbf{Boundary Loss. }
The illumination surrounding the shadow boundary of shadow images often exhibits abrupt variations, 
easily leading to boundary artifacts in the restored results.
In order to address this issue, a boundary-aware smoothness loss is intuitively employed to encourage smoother transitions along the boundaries in methods that require shadow masks.
Typically, the boundary-aware smoothness loss focuses on minimizing the gradients in the horizontal and vertical directions within the boundary regions during the training process~\cite{jin2021dc,guo2023boundary}, which is a type of non-reference loss and does not require ground truth shadow-free image as follows:
\begin{equation}
    \mathcal{L}_{\text {boundary }}=\left\|\mathrm{~B}\left(\mathbf{M}\right) \circ\left|\nabla\left(\hat{\mathbf{I}}_{sf}\right)\right|\right\|_1,
\end{equation}
where $\nabla$ is the gradient operation, $\mathrm{B}$ is a noise-robust function~\cite{xu2012structure,sharma2018into} to compute the boundaries of the shadow regions from the corresponding shadow mask $\mathbf{M}$.

\vspace{1mm}
\noindent\textbf{Adversarial Loss.}
Adversarial learning is a technique that involves solving a  maximization-minimization optimization problem~\cite{goodfellow2014generative} to foster enhanced results that are perceptually indistinguishable from reference images.
 It leverages the interplay between a generator and a discriminator to generate enhanced results that closely resemble the reference images, enabling the generation of high-quality and realistic outputs.

\vspace{1mm}
\noindent\textbf{Perceptual Loss.}
Perceptual loss is a technique used to encourage the generated results to closely resemble the ground truth in the feature space.
The loss enhances the visual quality of the outputs by considering high-level features.
It is computed as the Euclidean distance between the feature representations of an enhanced result and the corresponding ground truth. 
Typically, these feature representations are extracted from pre-trained networks, such as VGG~\cite{simonyan2015deep}, which enables the model to leverage the knowledge learned from a large dataset like ImageNet~\cite{deng2009imagenet}, making the perceptual loss more effective in capturing meaningful features.
\begin{equation}
    \mathcal{L}_{\text {perceptual }}=\|\phi(\mathbf{I}_{sf})-\phi(\hat{\mathbf{I}}_{sf})\|_2^2\;.
\end{equation}




\begin{table*}
	\centering
	\setlength\tabcolsep{5pt}
	\caption{
		{Summary of shadow datasets. `Syn' represents Synthetic.}
	}
	\label{tab:dataset}
  \adjustbox{width=1.\linewidth}{
	\begin{tabular}{r|c|c|c|c|c|c|c}
		\hline
		\textbf{Name}~~~~~~~~~~~~~~~~~~~~~&\textbf{Number}& \textbf{Resolution}&
		\textbf{Format} & \textbf{Real/Syn} & \textbf{Mask} & \textbf{Paired} & \textbf{Type} \\
		\hline
		Simulated by Physic Model~\cite{gryka2015learning,sidorov2019conditional,inoue2020learning} & +$\infty$ & $\infty$ &RGB 
		&Syn & &\\
		Simulated by GAN~\cite{le2019shadow, hu2019mask,cun2020towards} &+$\infty$  & $\infty$ &RGB 
		&Syn & &\\
		\hline
		UIUC~\cite{guo2012paired} & 108 &$640\times425$  &RGB 
		&Real  & \checkmark & \checkmark & soft, hard, self\\
		LRSS \cite{gryka2015learning} &137  &$5184\times3456$& raw 
		&Real & & \checkmark & soft \\
		SRD \cite{qu2017deshadownet}  &3,088  &$800\times 640$ &RGB 
		&Real  & & \checkmark & soft, hard \\
		ISTD \& ISTD+ \cite{wang2018stacked,le2019shadow}  & 1,870 &$640\times 480$ &RGB 
		&Real  & \checkmark & \checkmark & hard \\
  		USR \cite{hu2019mask}  &2,445  & $600\times 450$ &RGB 
		&Real  & &  & soft, hard \\
		SBU \cite{vicente2016large}  &5,000  & Various &RGB 
		&Real  & \checkmark & & soft, self, hard\\
		\hline
	\end{tabular}}
\end{table*}

\vspace{1mm}
\noindent\textbf{Color Loss.}
Some methods~\cite{wang2019underexposed,liu2021shadow} encourage the colour in restored result $\hat{\mathbf{I}}_{sf}$ to be the same with the shadow-free reference $\mathbf{I}_{sf}$ according to the extra colour loss as follows:
\begin{equation}
\begin{aligned}
\mathcal{L}_{\text {color }} =\sum_p\left(1-\cos <(\mathbf{I}_{sf}^{LAB})_p, (\hat{\mathbf{I}}_{sf}^{LAB})_p>\right),
\end{aligned}
\end{equation}
where $(\cdot)_p$ denotes the pixel with index $p$ and  and $\cos <\cdot, \cdot>$ represents
the cosine angle between vectors. 
Each pixel in the restored image or input image is treated as a 3D vector representing the Lab color space. The cosine similarity between two color vectors equals $1$ when the vectors are perfectly aligned.


\subsection{Training and Testing Datasets}

\vspace{1mm}
\noindent\textbf{UIUC~\cite{guo2012paired}}
comprises 108 images, each featuring a shadow image and its corresponding ground truth shadow-free image. Of these, 72 images are designated for training, and the remaining 76 for testing. The dataset includes various shadow types, such as soft shadows and self-shadows. Among the 76 test image pairs, 46 pairs are created by removing the shadow source while keeping the light source unchanged. In the remaining 30 pairs, shadows are cast by objects within the scene, with the shadow-free image generated by blocking the light source, enveloping the entire scene in shadow. The corresponding shadow mask is generated by thresholding the ratio between the paired shadow and shadow-free images.

\vspace{1mm}
\noindent\textbf{LRSS~\cite{gryka2015learning}} 
 is a collection of real soft shadow test photographs. It includes 137 images, 37 of which have corresponding ground truth shadow-free versions (along with mattes) to benchmark shadow removal methods. These ground truth images were produced by setting up a camera on a tripod and taking two photos: one with the shadow and one without, by removing the object casting the shadow.

\vspace{1mm}
\noindent\textbf{SRD~\cite{qu2017deshadownet}} comprises 2,680 training and 408 testing pairs of shadow and shadow-free images. This dataset does not feature manually annotated masks. Like ISTD, SRD includes shadows cast by various objects, 
then removing the shadow source to capture the corresponding shadow-free image. Shadow images are captured under different illumination conditions, such as varied weather conditions or different times of the day, to incorporate both hard and soft shadows into the dataset.

\vspace{1mm}
\noindent \textbf{ISTD and ISTD+~\cite{wang2018stacked,le2019shadow}} The ISTD dataset~\cite{wang2018stacked} comprises 1330 training and 540 testing triplets, consisting of shadow images, corresponding manually annotated masks, and shadow-free images. 
This dataset primarily comprises outdoor scenes and predominantly contains hard shadows. 
Shadows are cast by various objects not present in the scene, and to capture the corresponding shadow-free image, the shadow source is removed.
The Adjusted ISTD (ISTD+) dataset~\cite{le2019shadow} addresses illumination inconsistencies between the shadow and shadow-free images present in the original ISTD dataset.

\vspace{1mm}
\noindent\textbf{USR~\cite{hu2019mask}} 
is an unpaired shadow removal dataset consisting of 2,445 shadow images and 1,770 shadow-free images. It features a diverse range of scenes, encompassing over a thousand different environments where shadows are cast by various objects such as trees, buildings, traffic signs, people, umbrellas, railings, and more. The dataset includes 1,956 images for training and 489 for testing, utilizing all 1,770 shadow-free images exclusively for training.

\vspace{1mm}
\noindent\textbf{SBU~\cite{vicente2016large}}
 comprises 5,000 images depicting scenes with shadows, spanning diverse environments such as urban areas, beaches, mountains, roads, parks, snowscapes, animals, vehicles, and houses. The collection includes various types of photographs, including aerial shots, landscapes, close-ups, and selfies.
Approximately 85\% of the images are allocated to the training set, with the remaining 15\% reserved for testing. Among the 700 test images, detailed annotations of shadow masks ensure pixel accuracy. During training image annotation, a quick method involving sparse strokes on shadow areas was employed for efficient labeling of a large image set.


\vspace{1mm}
\noindent\textbf{Video~\cite{le2020shadow}}
 shadow removal dataset comprises 8 videos featuring static scenes with unchanging backgrounds. Each video includes a corresponding $V_{max}$ frame, which serves as the pseudo shadow-free reference, created by capturing the maximum intensity values at each pixel location throughout the video. The mask for moving shadows includes pixels found in both shadowed and non-shadowed areas, defining the evaluation region. The moving shadow mask is generated by applying a threshold of 80.


\begin{table*}[t]
	\renewcommand\arraystretch{1}
	\caption{Quantitative comparisons \bihan{of the shadow removal results} on SRD dataset~\cite{qu2017deshadownet} in terms of RMSE, PSNR (in dB), and SSIM. Highlighted in \colorbox{red!25}{red}, \colorbox{yellow!25}{yellow}, and {\colorbox[HTML]{CCECEB}{green}} cells are the best, second-best, and third-best results, respectively. $^\S$ indicates that the results are directly quoted from the original paper as \bihan{their code implementation} was not made available.}
	\vspace{-4pt}
	\begin{center}
 \adjustbox{width=1.\linewidth}{
		\setlength{\tabcolsep}{1.5mm}{
		\begin{tabular}{c|l|ccc|ccc|cccc}
			\hline
			\multirow{2}{*}{\textbf{Learning}} &\multirow{2}{*}{\textbf{Method}} & \multicolumn{3}{c|}{\textbf{Shadow Region (S)}} & \multicolumn{3}{c|}{\textbf{Non-Shadow Region (NS)}}  & \multicolumn{4}{c}{\textbf{All Image (ALL)}}\\
			& &	\textbf{PSNR}$\uparrow$ & \textbf{SSIM} $\uparrow$ & \textbf{RMSE}$\downarrow$ &	
\textbf{PSNR}$\uparrow$ & \textbf{SSIM} $\uparrow$ & \textbf{RMSE}$\downarrow$ & \textbf{PSNR}$\uparrow$ & \textbf{SSIM}$\uparrow$& \textbf{RMSE}$\downarrow$ &\textbf{LPIPS}$\downarrow$ 	 \\
			\hline
			 &input  &18.96 & 0.871 & 36.69 & 31.47 & 0.975 & 4.83 & 18.19 & 0.830 & 14.05 & 0.1899\\
			\hline
		 	TR & Guo \etal~\cite{guo2012paired}  & - & - & 29.89 & - & - & 6.47 & - & - & 12.60 & -\\ \hline
  	\multirow{11}*{SL}& DeshadowNet~\cite{qu2017deshadownet}&- &-&  11.78 & -&- & 4.84&-&- & 6.64 & -\\
		 	&DSC~\cite{hu2019direction}  & 30.65 & 0.960 & 8.62 & 31.94 & 0.965 & 4.41 & 27.76 & 0.903 & 5.71 & 0.1141\\
		 	&DHAN~\cite{cun2020towards} & 33.67 & 0.978 & 8.94 & 34.79 & 0.979 & 4.80 & 30.51 & 0.949 & 5.67 & 0.0666\\
		 	 &AEF~\cite{fu2021auto} & 32.26 & 0.966 & 9.55 & 31.87 & 0.945 & 5.74 & 28.40 & 0.893 & 6.50 & 0.0894\\
		 	&EMDN~\cite{zhu2022efficient} & 34.94 & 0.980 & 7.44 & 35.85 & 0.982 & 3.74 &31.72 & 0.952 & 4.79 & 0.0980\\
		 	&BMNet~\cite{zhu2022bijective}& 35.05 & 0.981 & 6.61 & 36.02 & 0.982 & 3.61 & 31.69 & 0.956 & 4.46 & 0.0549\\
   &ShadowFormer~\cite{guo2023shadowformer} &35.55 & 0.982 & 6.14 &  36.82 & 0.983 & 3.54 &32.46 &0.957 &4.28 & 0.0572\\
    	  &ShadowDiffusion~\cite{guo2022shadowdiffusion} &{\cellcolor{yellow!25}{38.72}} & {\cellcolor{red!25}{0.987}} & {\cellcolor{yellow!25}{4.98}}&  {\cellcolor{yellow!25}{37.78}}& {\cellcolor{yellow!25}{0.985}} & 3.44 &{\cellcolor{yellow!25}{34.73}} &{\cellcolor{yellow!25}{0.970}} &{\cellcolor[HTML]{CCECEB}{3.63}} & {\cellcolor{yellow!25}{0.0359}} \\

&Inpaint4Shadow~\cite{li2023leveraging} &36.73 & {\cellcolor[HTML]{CCECEB}{0.985}} & 5.70 &  36.70 & {\cellcolor{yellow!25}{0.985}} & {\cellcolor[HTML]{CCECEB}{3.27}} &33.27 &0.967 &3.81 & 0.0475\\
&DeS3~\cite{jin2022des3} & {\cellcolor[HTML]{CCECEB}{37.91}} & {\cellcolor{yellow!25}{0.986}} & {\cellcolor[HTML]{CCECEB}{5.27}} &  {\cellcolor[HTML]{CCECEB}{37.45}}& {\cellcolor[HTML]{CCECEB}{0.984}} & {\cellcolor{yellow!25}{3.03}} & {\cellcolor[HTML]{CCECEB}{34.11}} & {\cellcolor[HTML]{CCECEB}{0.968}} &{\cellcolor{yellow!25}{3.56}} & {\cellcolor{red!25}{0.0338}} \\
&HomoFormer~\cite{Xiao_2024_CVPR} & {\cellcolor{red!25}{38.81}} & {\cellcolor{red!25}{0.987}} & {\cellcolor{red!25}{4.25}}&  {\cellcolor{red!25}{39.45}} & {\cellcolor{red!25}{0.988}} & {\cellcolor{red!25}{2.85}} & {\cellcolor{red!25}{35.37}} & {\cellcolor{red!25}{0.972}} &{\cellcolor{red!25}{3.33}} & {\cellcolor[HTML]{CCECEB}{0.0419}}\\

			\hline
		 	UL	& DC-ShadowNet \cite{jin2021dc} & 34.00 & 0.975 & 7.70 & 35.53 & 0.981 & 3.65 & 31.53 & 0.955 & 4.65 & 0.1285\\
			\hline
	 		SSL	&ARGAN$^\S$ \cite{ding2019argan} & -& - &6.35 & - & - & 4.46 & - & - & 5.31 & -\\
			\hline
		\end{tabular}}}
	\end{center}
	\label{Table:srd}
\end{table*}

\begin{table*}[t]
 \renewcommand\arraystretch{1}
	\caption{Quantitative comparisons \bihan{of the shadow removal results} on ISTD+ dataset~\cite{wang2018stacked,le2019shadow} in terms of RMSE, PSNR (in dB), SSIM, and LPIPS. $^\dag$ indicates that the methods use the ground truth shadow mask provided by ISTD+ dataset as input.}
	\vspace{-4pt}
	\begin{center}
 \adjustbox{width=1.\linewidth}{
		\setlength{\tabcolsep}{1mm}{
		\begin{tabular}{c|l|ccc|ccc|cccc}
			\hline
			\multirow{2}{*}{\textbf{Learning}} &\multirow{2}{*}{\textbf{Method}} & \multicolumn{3}{c|}{\textbf{Shadow Region (S)}} & \multicolumn{3}{c|}{\textbf{Non-Shadow Region (NS)}}  & \multicolumn{4}{c}{\textbf{All Image (ALL)}}\\
			& &	\textbf{PSNR}$\uparrow$ & \textbf{SSIM} $\uparrow$ & \textbf{RMSE}$\downarrow$ &	\textbf{PSNR}$\uparrow$ & \textbf{SSIM} $\uparrow$ & \textbf{RMSE}$\downarrow$ & \textbf{PSNR}$\uparrow$ & \textbf{SSIM}$\uparrow$& \textbf{RMSE}$\downarrow$& \textbf{LPIPS}$\downarrow$   \\
			\hline
			 &input  & 20.83 & 0.93 & 39.01 & 37.46 & 0.985 & 2.40 & 20.46 & 0.894 & 8.40 & 0.1431\\
			\hline
		 	TR & Guo \etal~\cite{guo2012paired}  & 26.89 & 0.960 & 21.38 & 35.56 & 0.976 & 3.09 & 25.52 & 0.925 & 6.09 & 0.0966 \\ \hline
		 	\multirow{9}*{SL}&SP+M-Net~\cite{le2019shadow} & 37.60 & 0.990 & 5.91 & 36.02 & 0.976 & 2.99 & 32.94 & 0.962 & 3.46 & 0.0586\\
    &Param+M+D-Net~\cite{le2020shadow} & 33.09 & 0.983 & 9.67 & 35.35 & 0.978 & 2.82 & 30.15 & 0.951 & 3.94 & 0.0662\\
		 	 &AEF~\cite{fu2021auto} & 36.04 & 0.978 & 6.55 & 31.16 & 0.892 & 3.77 & 29.45 & 0.861 & 4.23 & 0.0615\\

   		 	&BMNet$^\dag$~\cite{zhu2022bijective}& 37.87 & {\cellcolor[HTML]{CCECEB}{0.991}} & 5.62 & 37.51 & 0.985 & 2.45 & 33.98 & {\cellcolor{yellow!25}{0.972}} & 2.97 & 0.0305\\
       &SG-ShadowNet~\cite{wan2022style} & 36.80 & 0.990 & 5.93 & 35.57 & 0.978 & 2.92 & 32.46 & 0.962 & 3.41 & 0.0430\\
        &Inpaint4Shadow~\cite{li2023leveraging}&38.10 & 0.990 & 6.09 & 37.66 & 0.981 & 2.82 &  34.16 & 0.967 & 3.35 & 0.0675\\
&ShadowFormer$^\dag$~\cite{guo2023shadowformer}& {\cellcolor[HTML]{CCECEB}{39.48}} & {\cellcolor{yellow!25}{0.992}} & {\cellcolor[HTML]{CCECEB}{5.23}} & {\cellcolor{yellow!25}{38.82}} & {\cellcolor[HTML]{CCECEB}{0.983}} & {\cellcolor[HTML]{CCECEB}{2.30}} & {\cellcolor{yellow!25}{35.46}} & {\cellcolor[HTML]{CCECEB}{0.971}} & {\cellcolor[HTML]{CCECEB}{2.78}} & {\cellcolor[HTML]{CCECEB}{0.0260}}\\
&ShadowDiffusion$^\dag$~\cite{guo2022shadowdiffusion}&{\cellcolor{red!25}{39.69}} & {\cellcolor{yellow!25}{0.992}} & {\cellcolor{yellow!25}{4.97}} & {\cellcolor{red!25}{38.89}} & {\cellcolor{red!25}{0.987}} & {\cellcolor{yellow!25}{2.28}} & {\cellcolor{red!25}{35.67}} & {\cellcolor{red!25}{0.975}} & {\cellcolor{yellow!25}{2.72}} & {\cellcolor{red!25}{0.0234}}\\
&HomoFormer$^\dag$~\cite{Xiao_2024_CVPR} &{\cellcolor{yellow!25}{39.49}} &  {\cellcolor{red!25}{0.993}} & {\cellcolor{red!25}{4.73}}&  {\cellcolor[HTML]{CCECEB}{38.75}} & {\cellcolor{yellow!25}{0.984}}  & {\cellcolor{red!25}{2.23}} & {\cellcolor[HTML]{CCECEB}{35.35}} & {\cellcolor{red!25}{0.975}} & {\cellcolor{red!25}{2.64}} & {\cellcolor{yellow!25}{0.0259}}\\
			\hline
		 	UL	& DC-ShadowNet \cite{jin2021dc} & 31.06 & 0.976 & 12.62 & 27.03 & 0.961 & 6.82 & 25.03 & 0.926 & 7.77 & 0.1420\\
			\hline
		 	\multirow{2}*{ZSL}&	G2R-ShadowNet \cite{liu2021from} & 33.58& 0.979 & 8.82 & 35.52 &0.976& 2.89 & 30.52 & 0.944 & 3.86 & 0.0662\\
		 		&BCDiff \cite{guo2023boundary} &35.71 & 0.986 & 7.61&  36.39 & 0.981 & 2.66&32.11 &0.959 &3.47 & 0.0518\\			
			\hline
		\end{tabular}}}
	\end{center}
	\label{Table:istd+}
\end{table*}

\begin{table*}[t]
 \renewcommand\arraystretch{1}
	\caption{Quantitative comparisons on LRSS dataset~\cite{gryka2015learning} and UIUC~\cite{guo2012paired} in terms of RMSE, PSNR (in dB), SSIM, and LPIPS.}
	\vspace{-4pt}
	\begin{center}
		 \adjustbox{width=.85\linewidth}{
		\begin{tabular}{c|l|ccc|ccc}
			\hline
			\multirow{2}{*}{\textbf{Learning}} &\multirow{2}{*}{\textbf{Method}} & \multicolumn{3}{c|}{\textbf{LRSS}} & \multicolumn{3}{c}{\textbf{UIUC}}\\
			& &	\textbf{PSNR}$\uparrow$ & \textbf{SSIM} $\uparrow$ & \textbf{RMSE}$\downarrow$  &	\textbf{PSNR}$\uparrow$ & \textbf{SSIM} $\uparrow$ & \textbf{RMSE}$\downarrow$ \\
			\hline
			 &input  & 16.43 &0.886 &15.98 &20.85 & 0.803& 13.97\\
			\hline
		 	&SP+M-Net~\cite{le2019shadow} & 21.77 & 0.927 & 11.18& 28.23& 0.866 & 7.13 \\
          &EMDN~\cite{zhu2022efficient}&19.26 &0.882 & 15.71& 25.56 & 0.775 & 13.07 \\
     &BMNet~\cite{zhu2022bijective}& 14.19 &0.048 & 50.79 & 16.86 & 0.679 & 45.68 \\
        SL &Inpaint4Shadow~\cite{li2023leveraging}&19.59 & 0.805 & 13.79& 27.77 & 0.849 & 7.82\\
&ShadowFormer~\cite{guo2023shadowformer}& {\cellcolor{yellow!25}{27.01}} & {\cellcolor{red!25}{0.957}} & {\cellcolor{yellow!25}{6.37}} & {\cellcolor[HTML]{CCECEB}{28.96}} & {\cellcolor{yellow!25}{0.874}} & {\cellcolor{red!25}{6.57}} \\
&ShadowDiffusion~\cite{guo2022shadowdiffusion}& {\cellcolor{red!25}{28.37}}  & {\cellcolor{yellow!25}{0.955}}  & {\cellcolor{red!25}{6.01}}& {\cellcolor{yellow!25}{29.02}} & {\cellcolor{red!25}{0.880}} & {\cellcolor[HTML]{CCECEB}{7.05}}  \\
&HomoFormer~\cite{Xiao_2024_CVPR} & {\cellcolor[HTML]{CCECEB}{25.86}} & {\cellcolor[HTML]{CCECEB}{0.951}} & {\cellcolor[HTML]{CCECEB}{6.98}} & {\cellcolor{red!25}{29.08}} & {\cellcolor[HTML]{CCECEB}{0.870}} & {\cellcolor{yellow!25}{6.81}} \\
			\hline
		 	UL	& DC-ShadowNet \cite{jin2021dc} & 20.89 & 0.902 &12.55 & 24.85 & 0.849 & 9.51\\
			\hline
		 	\multirow{2}*{ZSL}&	G2R-ShadowNet \cite{liu2021from} & 20.90 &0.901 & 9.99& 27.56 & 0.858 & 7.43\\
		 		&BCDiff \cite{guo2023boundary} & 22.13 & 0.922 & 10.68 & 26.81 & 0.852 & 7.96 \\			
			\hline
		\end{tabular}}
	\end{center}
	\label{Table:lrss_uiuc}
\end{table*}

\subsection{Evaluation Metrics}

\begin{itemize}
    \item \textbf{PSNR and RMSE.} 
PSNR measures the ratio between the maximum signal power and the power of noise affecting image fidelity, while RMSE calculates the square root of the average squared difference between corresponding pixels of the reference and reconstructed images. Both metrics quantify the discrepancy between images, with lower values indicating higher distortion or error.

   \item \textbf{MAE} measures the average magnitude of errors between predicted shadow-free image and ground truth. It calculates the absolute difference between each predicted value and its corresponding ground truth value, then averages these absolute differences to provide a single score.

\item \textbf{SSIM}~\cite{wang2004image} is a metric evaluating similarity between images, considering perceptual differences, and treating image degradation as a perceived alteration in structural information. It measures luminance, contrast, and structure, providing a score between -1 and 1, with 1 indicating perfect structural similarity.

\item \textbf{LPIPS}~\cite{zhang2018perceptual} is a perceptual similarity metric that evaluates differences between images based on their perceived similarity by a deep neural network. It measures the distance between feature representations of images, capturing both low-level and high-level features. Lower LPIPS scores indicate greater perceptual similarity, making it effective for tasks where human judgment is crucial, like image generation and restoration.

\item \textbf{NIQE}~\cite{mittal2012making} is a no-reference metric for estimating image quality. It analyzes local statistics like luminance and contrast, comparing them to those of natural images to compute a quality score. Higher NIQE scores indicate lower image quality. It's useful for tasks like compression and denoising but may not perform as well on structured or synthetic images.

\end{itemize}

\section{Benchmarking and Empirical Analysis}\label{sec:benchmark}

This section provides empirical analysis and highlights some key challenges in deep learning based shadow removal methods.
We conduct extensive evaluations on several public benchmarks.
We select a number of recent shadow removal algorithms from different categories to be evaluated:

\begin{enumerate}
    \item Multi-context Embedding, DeShadowNet~\cite{qu2017deshadownet}

    \item Attentive Recurrent GAN, ARGAN~\cite{ding2019argan}

    \item Direction-aware Spatial Context, DSC~\cite{hu2019direction}

    \item Shadow Parameter Estimation and Matte Prediction Network, SP+M-Net~\cite{le2019shadow}

    \item Dual Hierarchical Aggregation Network, DHAN~\cite{cun2020towards}

    \item Generation to Removal, G2R-ShadowNet~\cite{liu2021from}

    \item Domain Classifier Guided Net, DC-ShadowNet~\cite{jin2021dc}

    \item Auto-Exposure Fusion, AEF~\cite{fu2021auto}

    \item Efficient Model-Driven Network, EMDN~\cite{zhu2022efficient}

    \item Bijective Mapping Network, BMNet~\cite{zhu2022bijective}
    
    \item Context Transformer, ShadowFormer~\cite{guo2023shadowformer}
    
    \item Unrolling-inspired Diffusion, ShadowDiffusion~\cite{guo2022shadowdiffusion}

    \item Boundary-aware Conditional Diffusion, BCDiff~\cite{guo2023boundary}

      \item Diffusion-based Soft \& Self Shadow Removal, DeS3~\cite{jin2022des3}
   \item Homogenized Transformer, HomoFormer~\cite{Xiao_2024_CVPR}
\end{enumerate}

    including eleven supervised learning-based methods (DeShadowNet, DSC, SP+M-Net, DHAN, AEF, EMDN, BMNet, ShadowFormer, ShadowDiffusion, SeS3, HomoFormer), two zero-shot learning-based methods (G2R-ShadowNet, BCDiff), one unsupervised learning-based method (DC-ShadowNet), one semi-supervised learning-based method (ARGAN). To ensure fair comparisons, we utilize publicly available code for all the methods under consideration to generate their results.
    We leverage full-reference metrics including PSNR, SSIM, RMSE, and LPIPS for those testsets with ground truth shadow-free images, \ie, ISTD+~\cite{wang2018stacked,le2019shadow}, SRD~\cite{qu2017deshadownet}, LRSS~\cite{gryka2015learning}, and UIUC~\cite{guo2012paired}.
    The results of shadow removal from all methods are either sourced from their original papers or replicated using their official implementations. 
    We standardize the evaluation of shadow removal results to a resolution of $256\times 256$ following most shadow removal methods~\cite{fu2021auto,zhu2022efficient,zhu2022bijective,guo2022shadowdiffusion}.

\begin{figure*}[tbp]
    \centering
    \vspace{-3mm}
    \subfigure[Input]{
        \includegraphics[width=0.19\linewidth, height=0.19\linewidth]{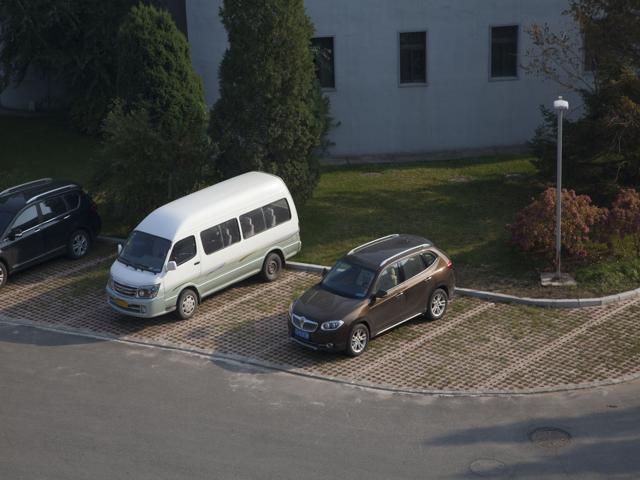}
    }\hspace{-6.5pt}
    \subfigure[Mask (Residual)]{ 
        \includegraphics[width=0.19\linewidth, height=0.19\linewidth]{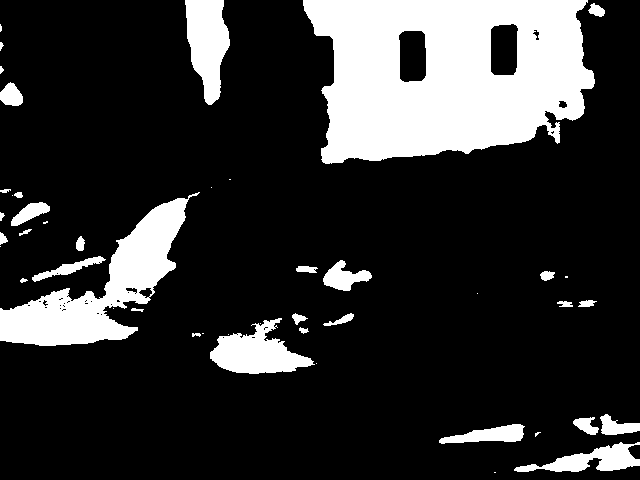}
    }\hspace{-6.5pt}
    \subfigure[Mask (DHAN)]{ 
        \includegraphics[width=0.19\linewidth, height=0.19\linewidth ]{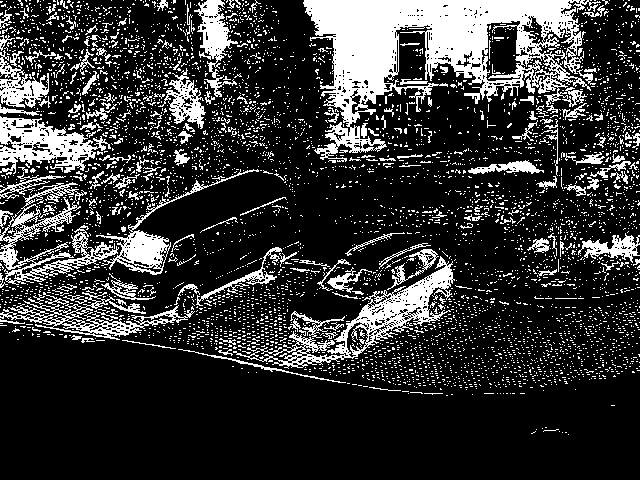}
    }\hspace{-6.5pt}
    \subfigure[DSC]{ 
        \includegraphics[width=0.19\linewidth, height=0.19\linewidth]{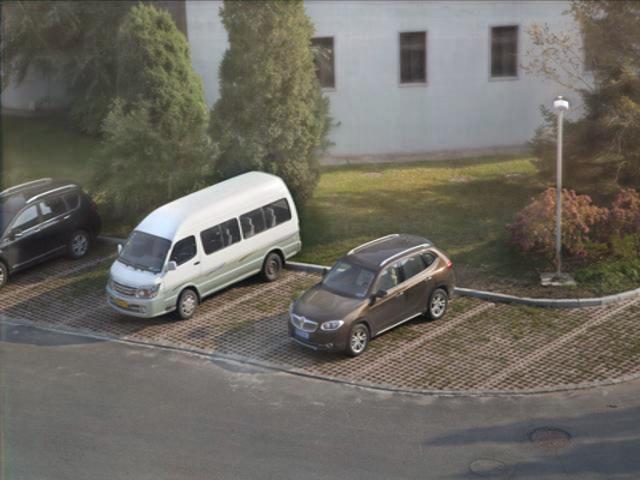}
    }\hspace{-6.5pt}
    \subfigure[DHAN]{ 
        \includegraphics[width=0.19\linewidth, height=0.19\linewidth]{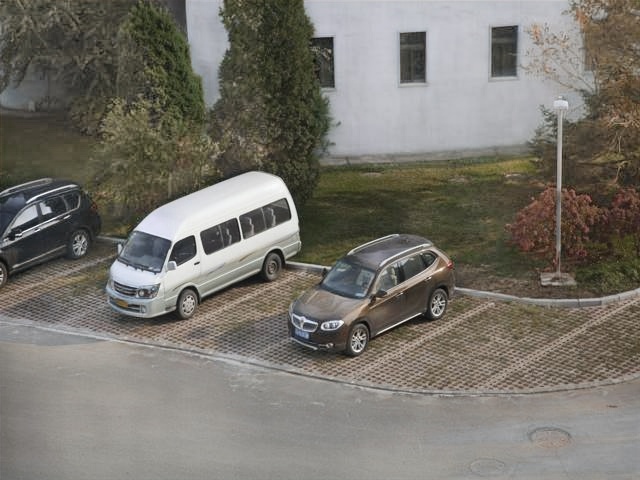}
    }
    \\
    \subfigure[AEF]{ 
        \includegraphics[width=0.19\linewidth, height=0.19\linewidth]{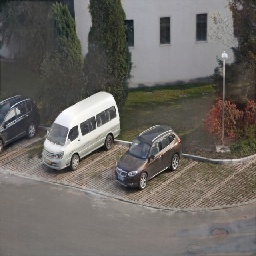}
    }\hspace{-6.5pt}
    \subfigure[EMDN]{ 
        \includegraphics[width=0.19\linewidth, height=0.19\linewidth]{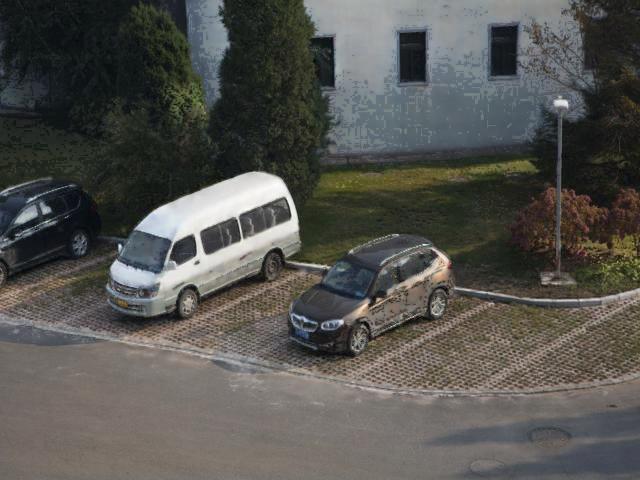}
    }\hspace{-6.5pt}
    \subfigure[BMNet]{ 
        \includegraphics[width=0.19\linewidth, height=0.19\linewidth]{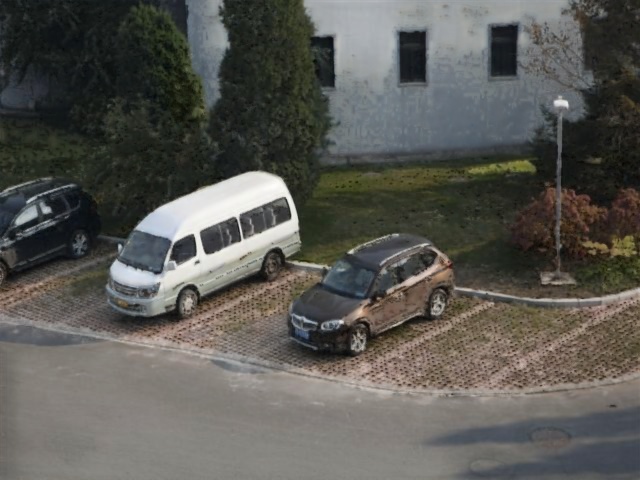}
    }\hspace{-6.5pt}
    \subfigure[DC-ShadowNet]{
        \includegraphics[width=0.19\linewidth, height=0.19\linewidth]{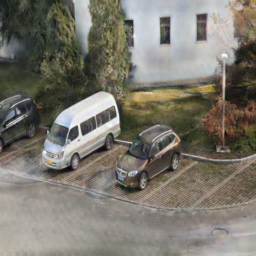}
    }\hspace{-6.5pt}
    \subfigure[SG-ShadowNet]{
        \includegraphics[width=0.19\linewidth, height=0.19\linewidth]{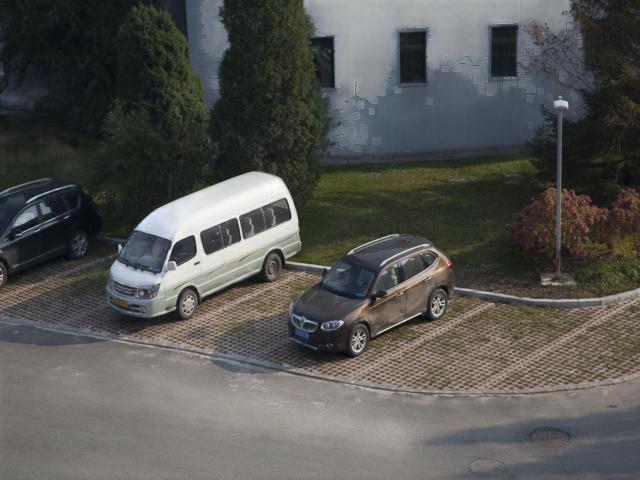}
    }
        \\
    \subfigure[Inpaint4Shadow]{ 
        \includegraphics[width=0.19\linewidth, height=0.19\linewidth]{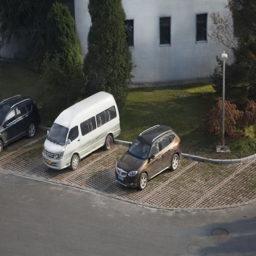}
    }\hspace{-6.5pt}
    \subfigure[ShadowDiffusion]{ 
        \includegraphics[width=0.19\linewidth, height=0.19\linewidth]{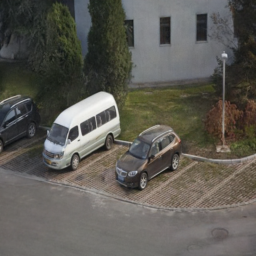}
    }\hspace{-6.5pt}
    \subfigure[DeS3]{ 
        \includegraphics[width=0.19\linewidth, height=0.19\linewidth]{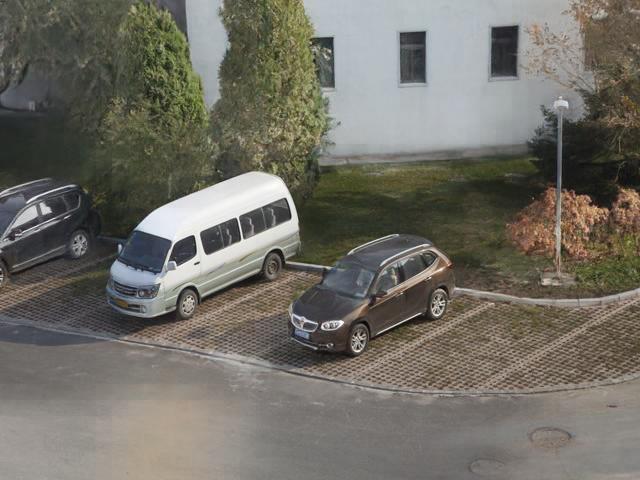}
    }\hspace{-6.5pt}
    \subfigure[HomoFormer]{
        \includegraphics[width=0.19\linewidth, height=0.19\linewidth]{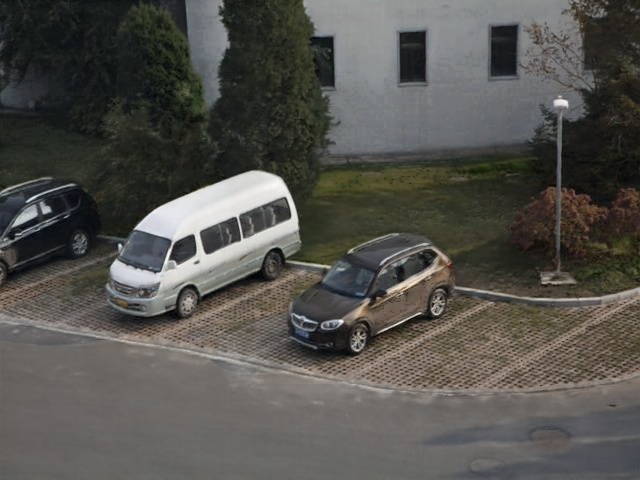}
    }\hspace{-6.5pt}
    \subfigure[Reference]{
        \includegraphics[width=0.19\linewidth, height=0.19\linewidth]{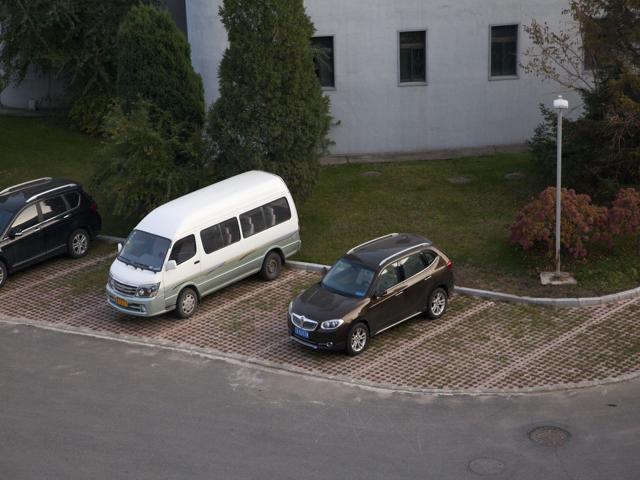}
    }
    \caption{Visual comparison with different shadow removal methods sampled from test set of SRD~\cite{qu2017deshadownet}.}
    \label{fig:srd}
\end{figure*}

\subsection{Benchmarking Results}

\begin{figure*}[tbp]
    \centering
    \vspace{-3mm}
    \subfigure[Input]{
        \includegraphics[width=0.18\linewidth]{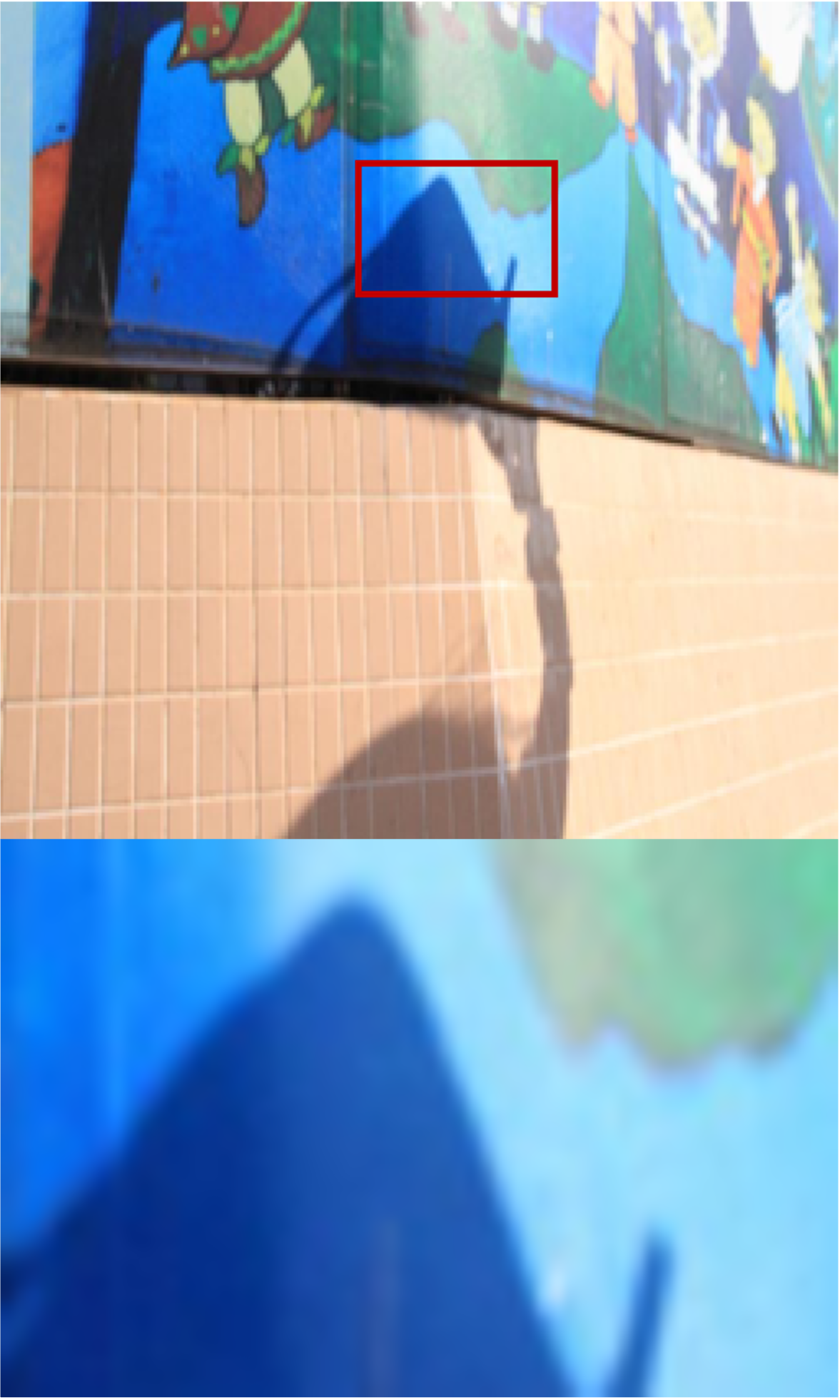}
    }\hspace{-6.5pt}
    \subfigure[Mask (Residual)]{ 
        \includegraphics[width=0.18\linewidth]{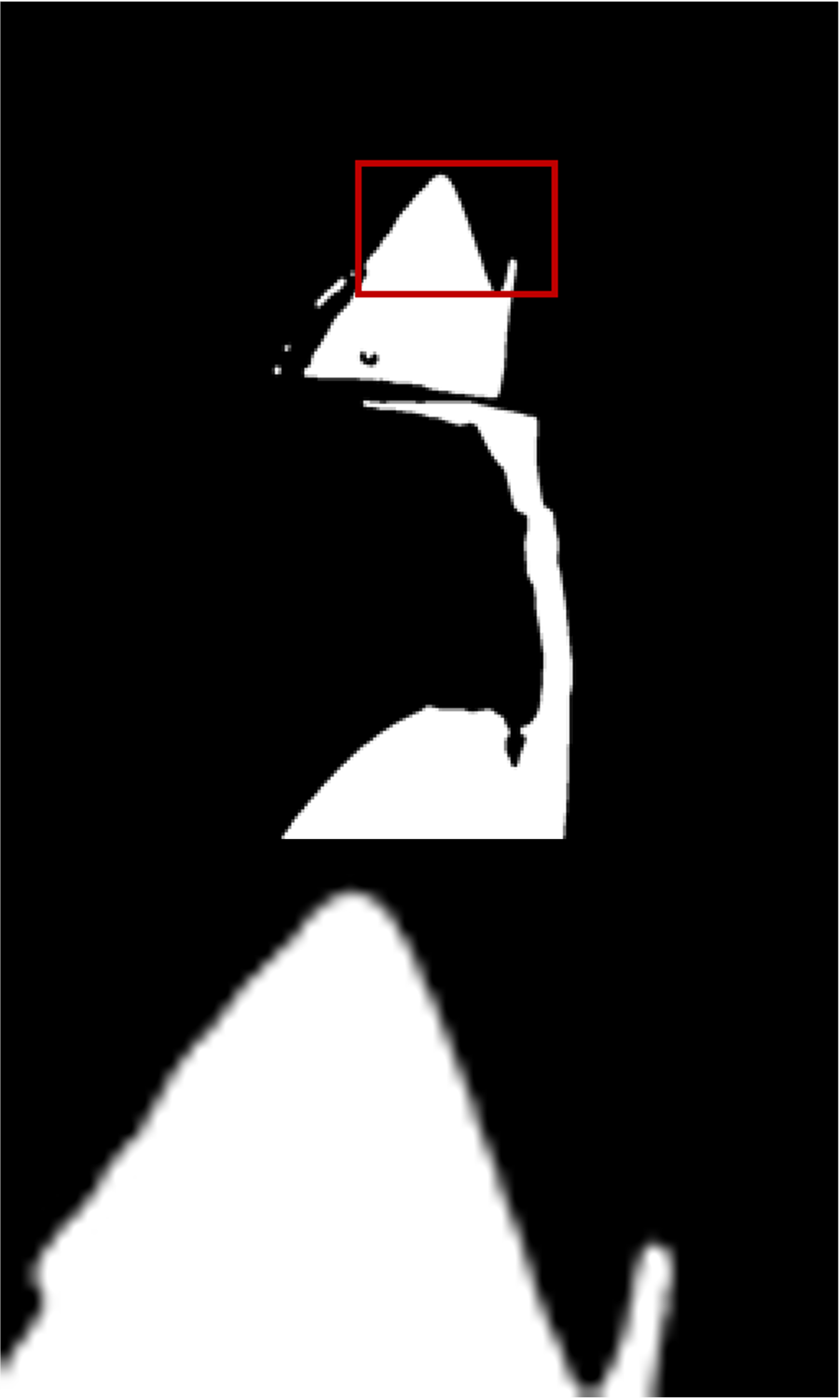}
    }\hspace{-6.5pt}
    \subfigure[Mask (DHAN)]{ 
        \includegraphics[width=0.18\linewidth]{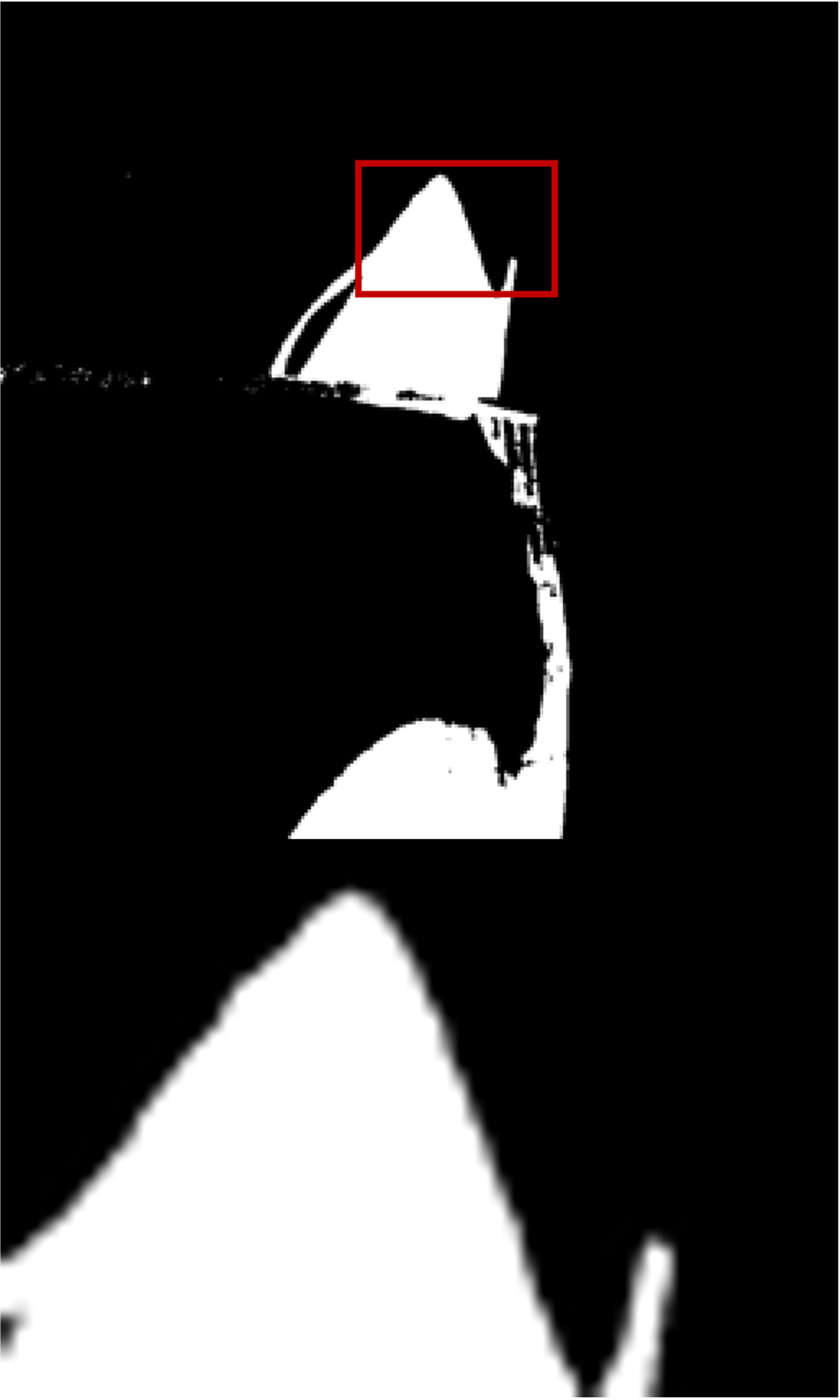}
    }\hspace{-6.5pt}
    \subfigure[DSC]{ 
        \includegraphics[width=0.18\linewidth]{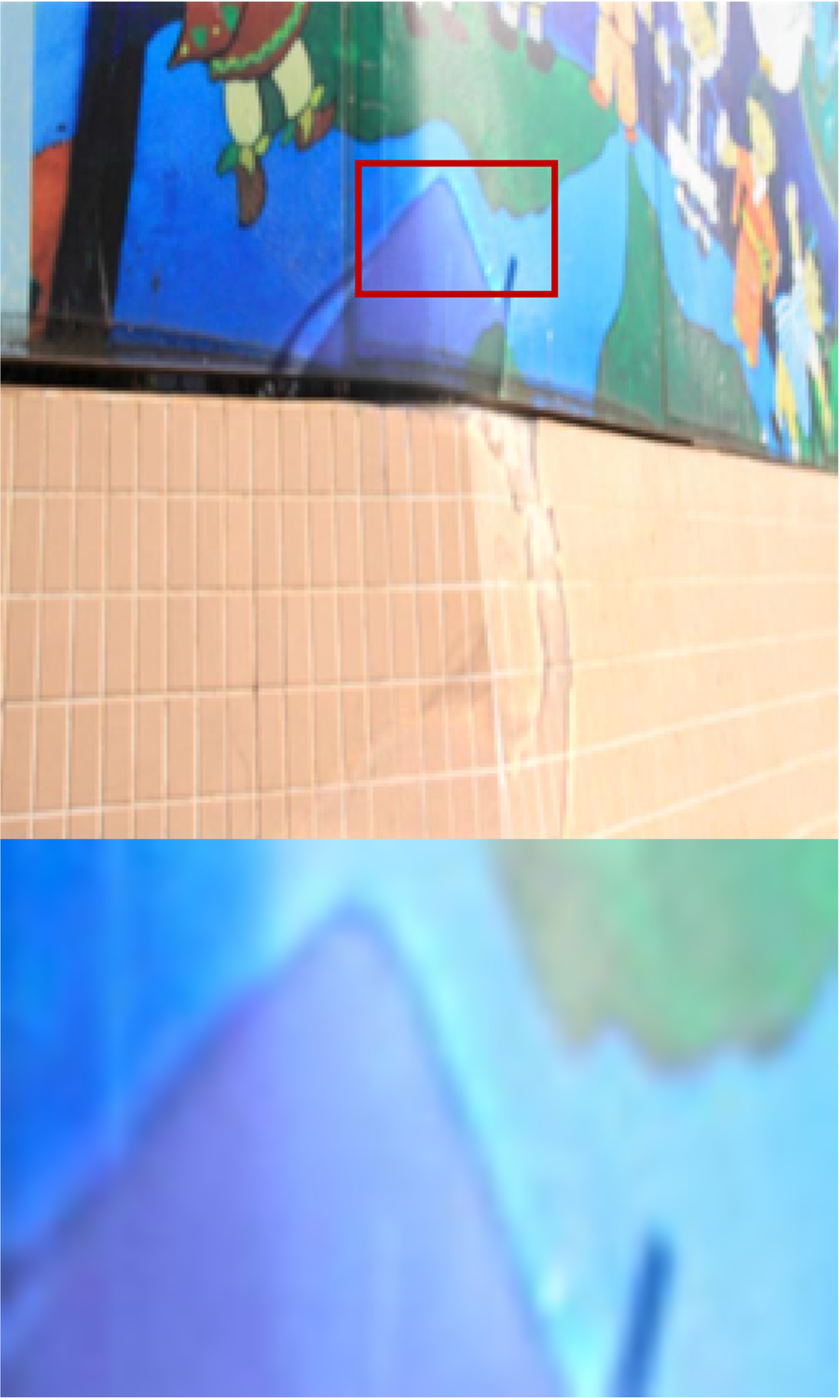}
    }\hspace{-6.5pt}
    \subfigure[DHAN]{ 
        \includegraphics[width=0.18\linewidth]{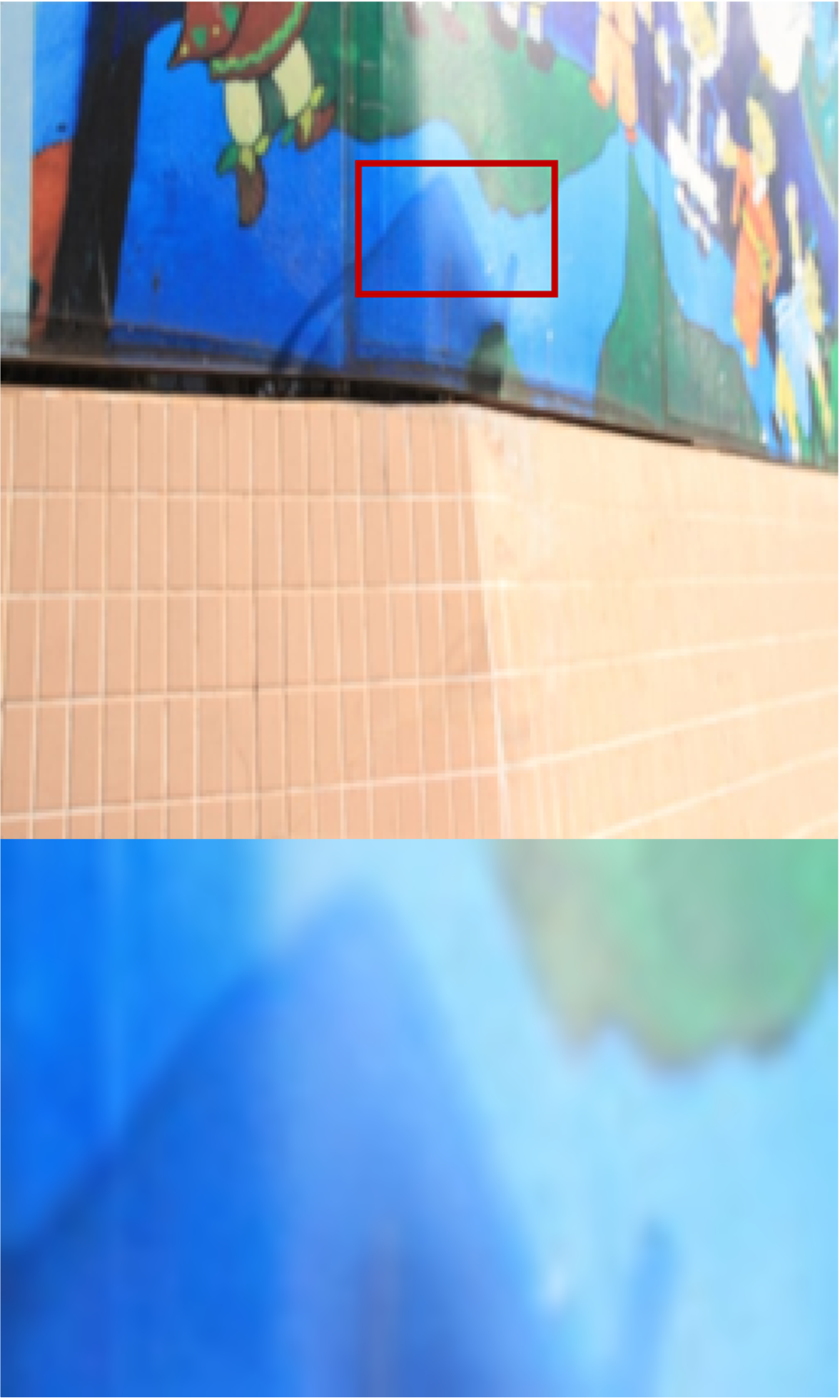}
    }
    \\
    \subfigure[AEF]{ 
        \includegraphics[width=0.18\linewidth]{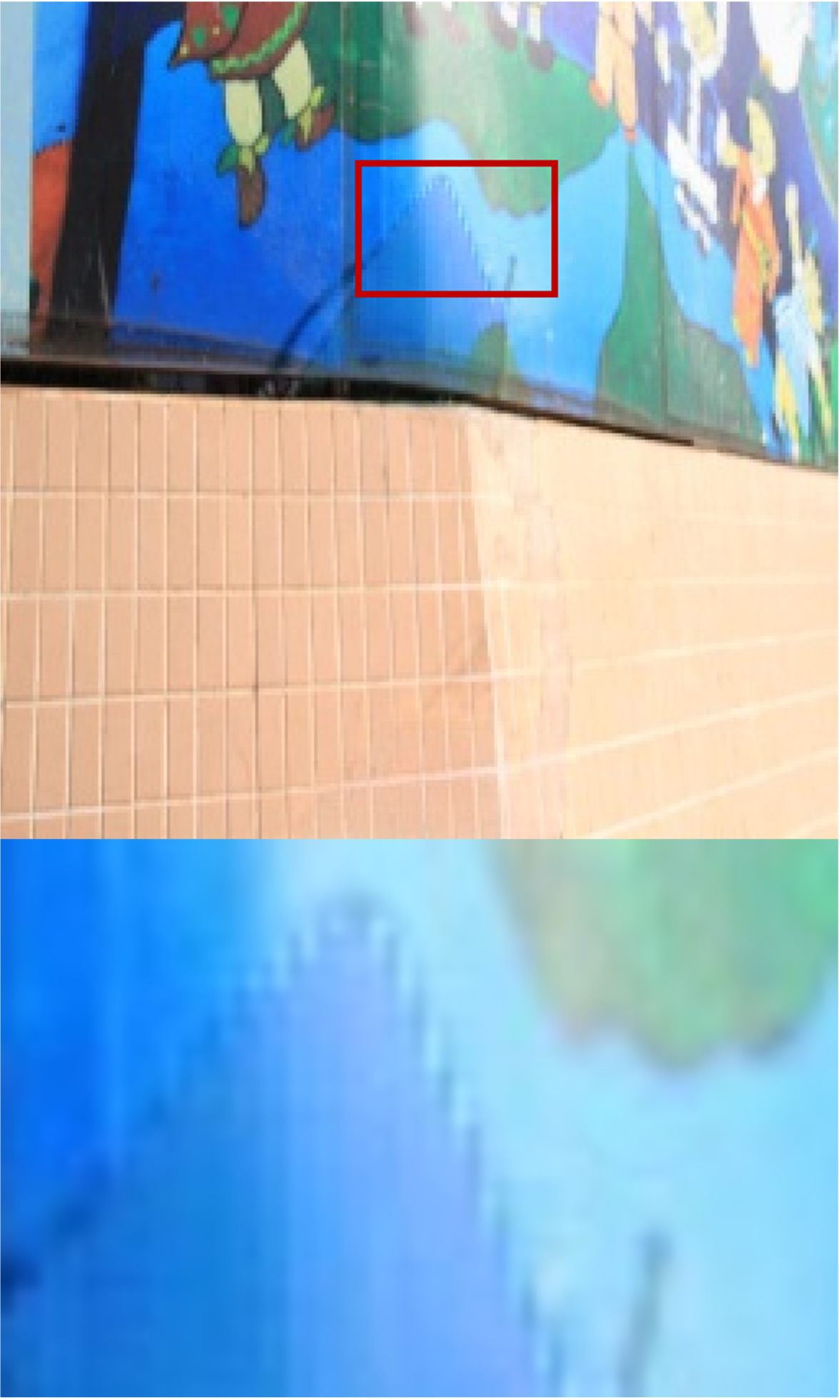}
    }\hspace{-6.5pt}
    \subfigure[EMDN]{ 
        \includegraphics[width=0.18\linewidth]{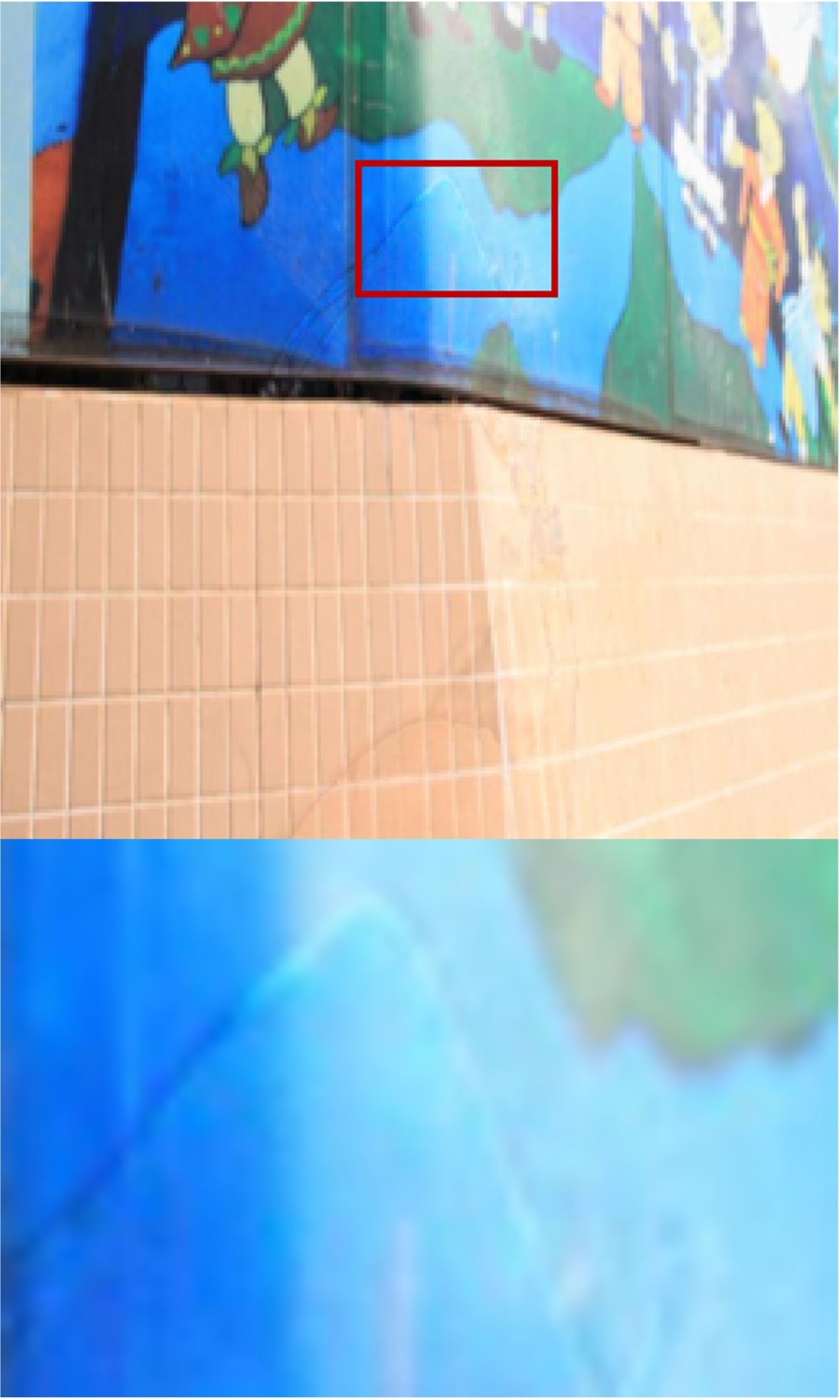}
    }\hspace{-6.5pt}
    \subfigure[BMNet]{ 
        \includegraphics[width=0.18\linewidth]{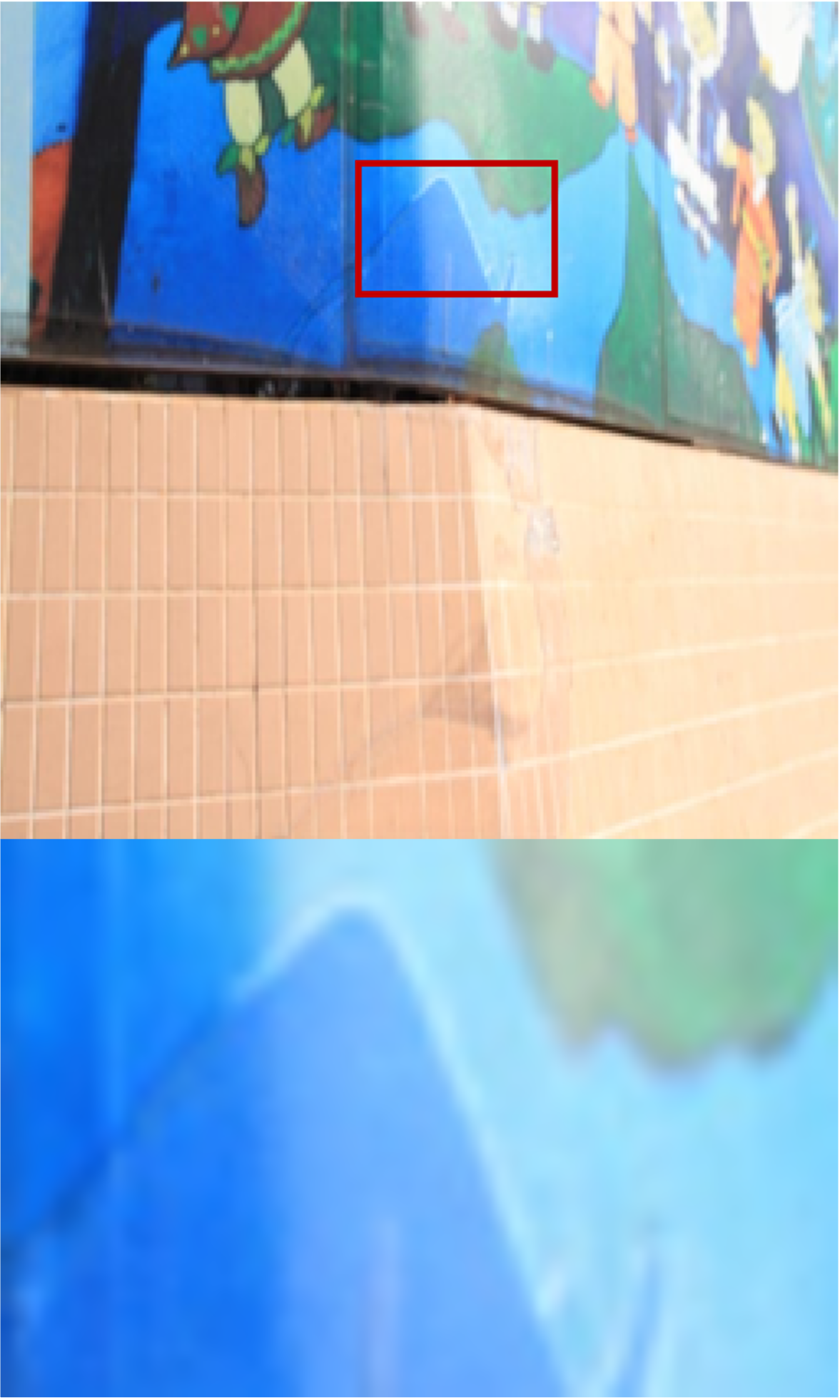}
    }\hspace{-6.5pt}
    \subfigure[DC-ShadowNet]{
        \includegraphics[width=0.18\linewidth]{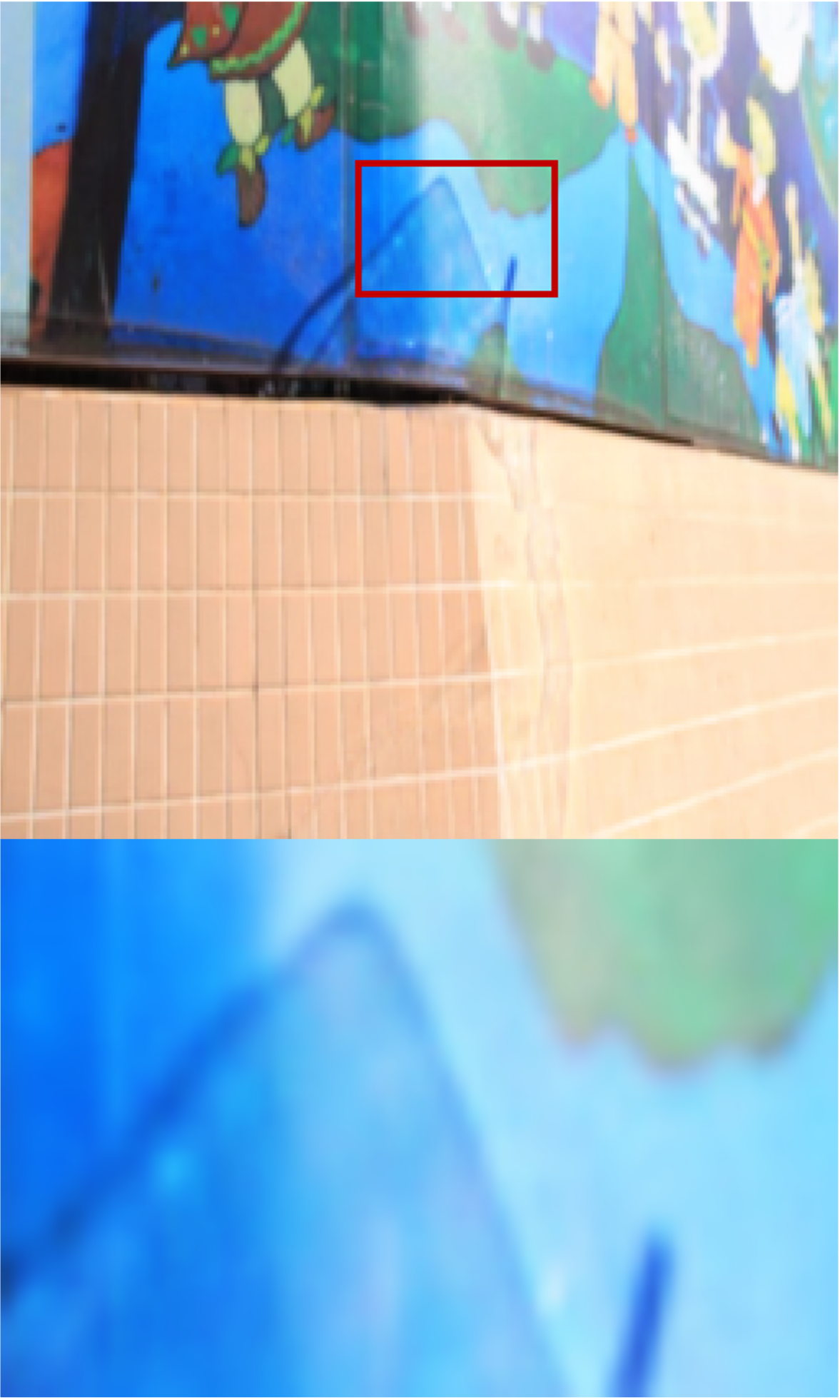}
    }\hspace{-6.5pt}
    \subfigure[SG-ShadowNet]{
        \includegraphics[width=0.18\linewidth]{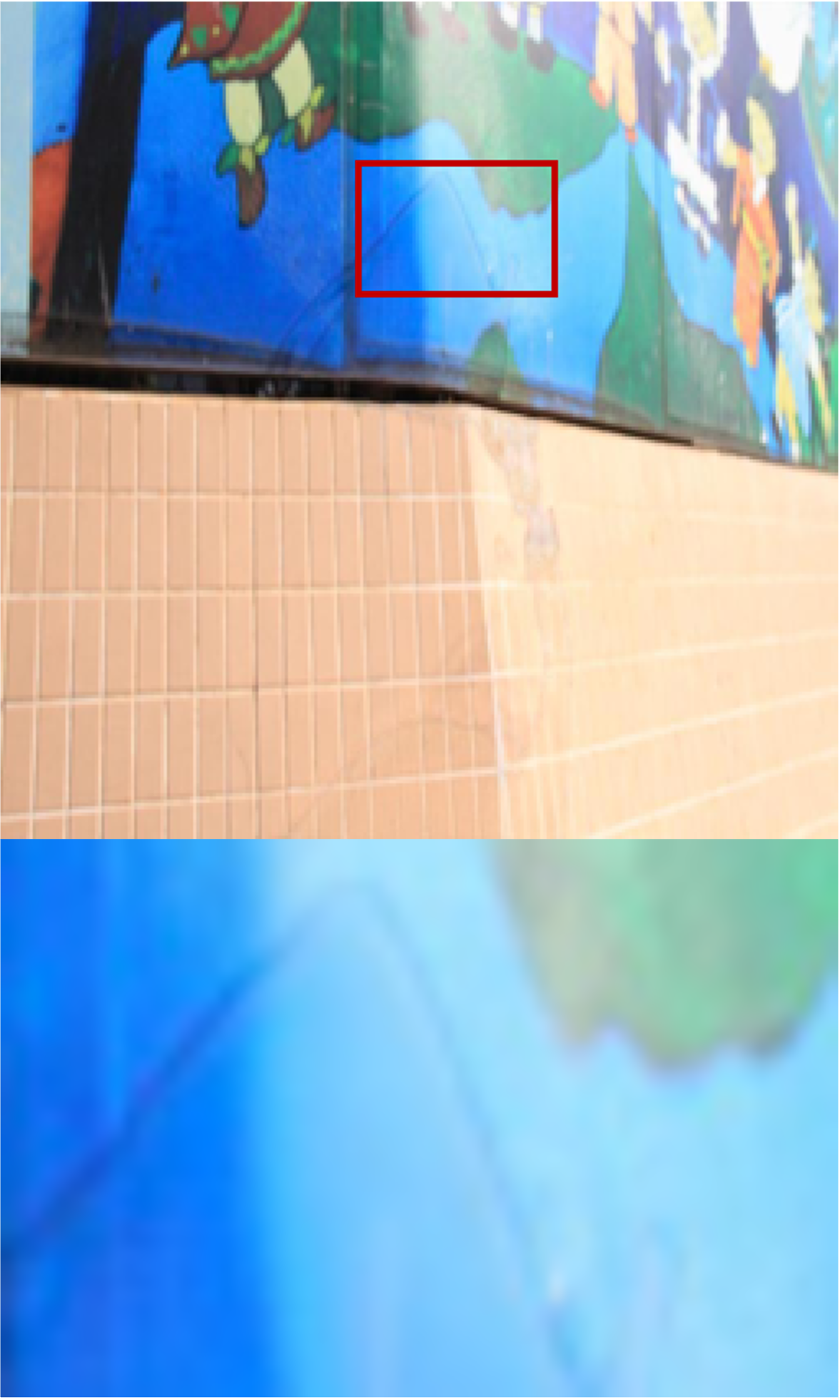}
    }
        \\
    \subfigure[Inpaint4Shadow]{ 
        \includegraphics[width=0.18\linewidth]{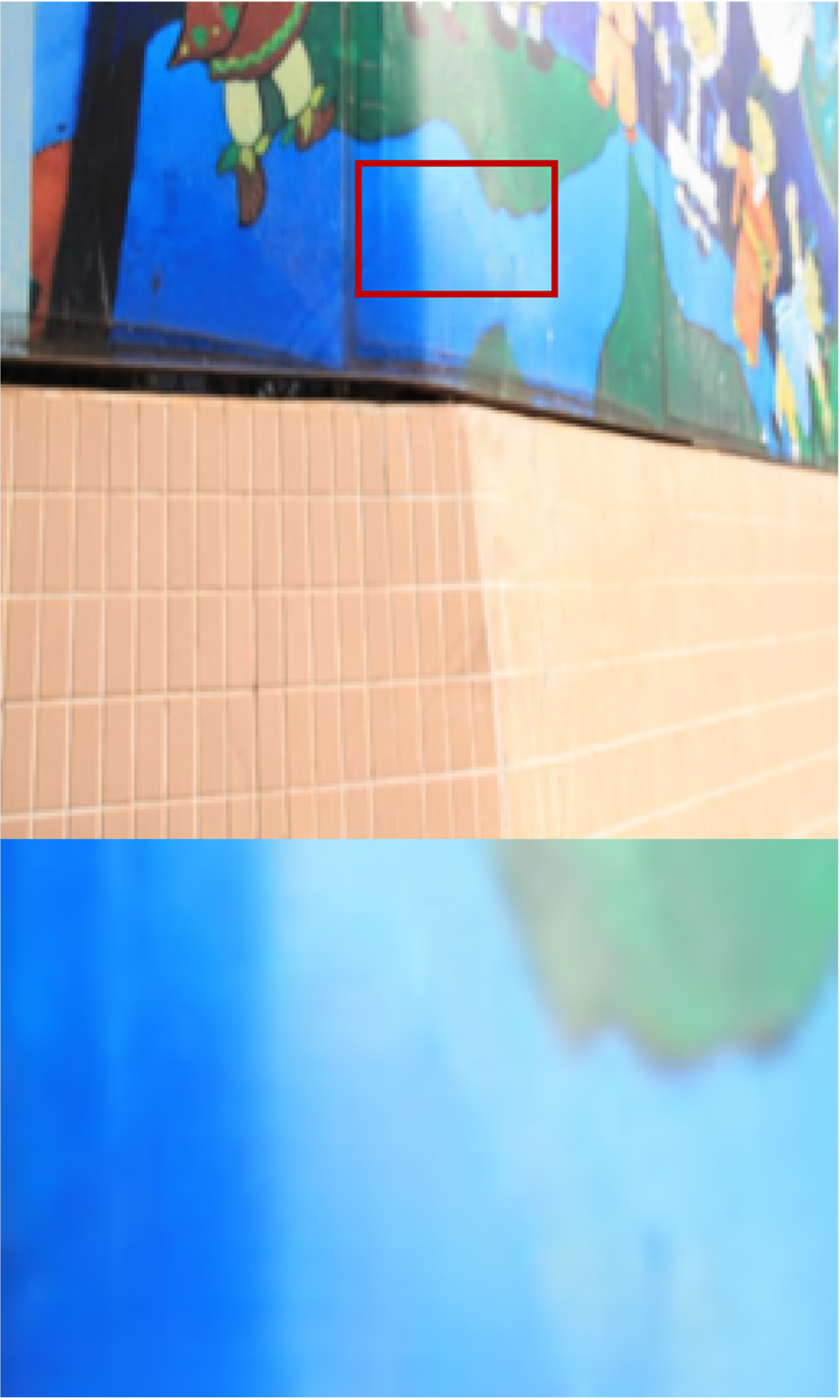}
    }\hspace{-6.5pt}
    \subfigure[ShadowDiffusion]{ 
        \includegraphics[width=0.18\linewidth]{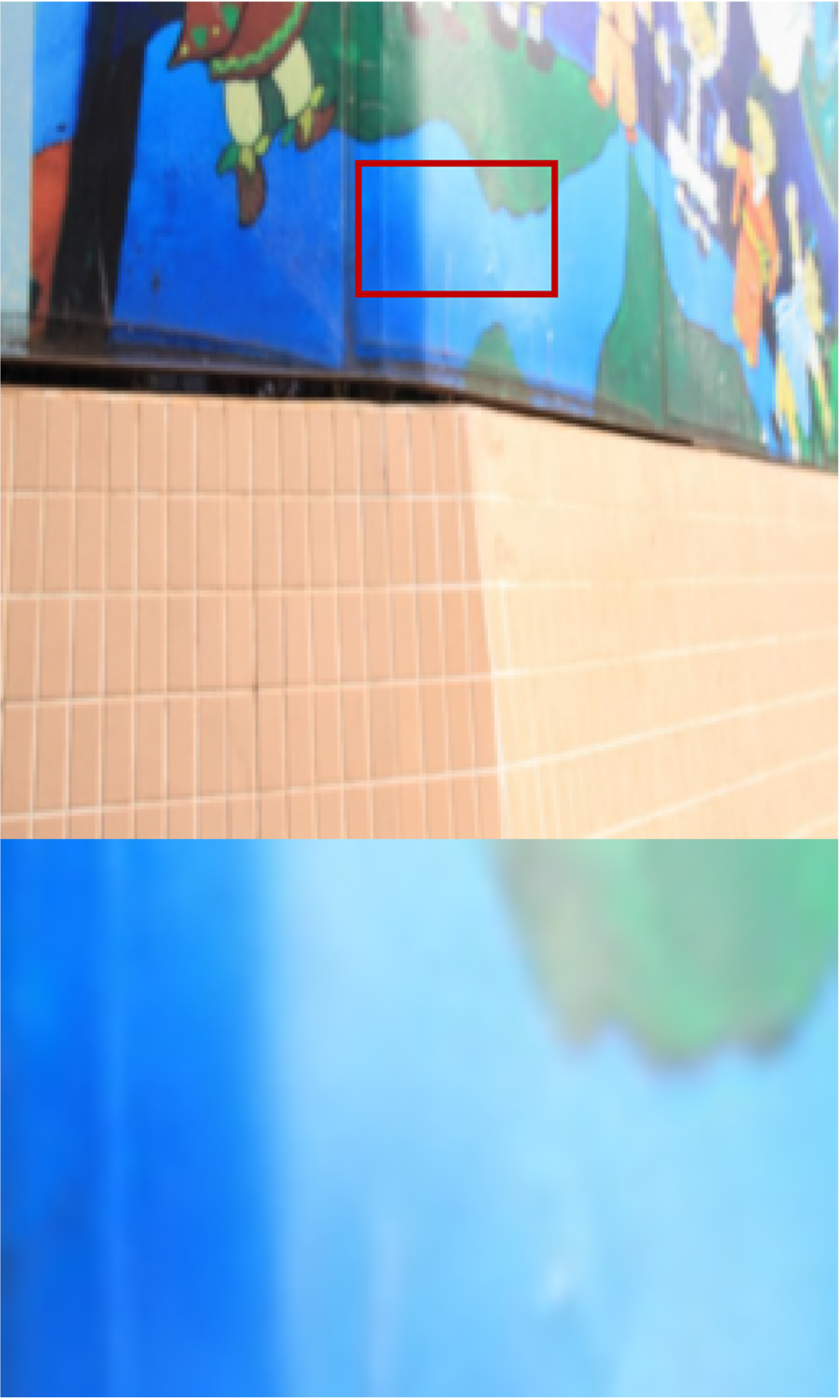}
    }\hspace{-6.5pt}
    \subfigure[DeS3]{ 
        \includegraphics[width=0.18\linewidth]{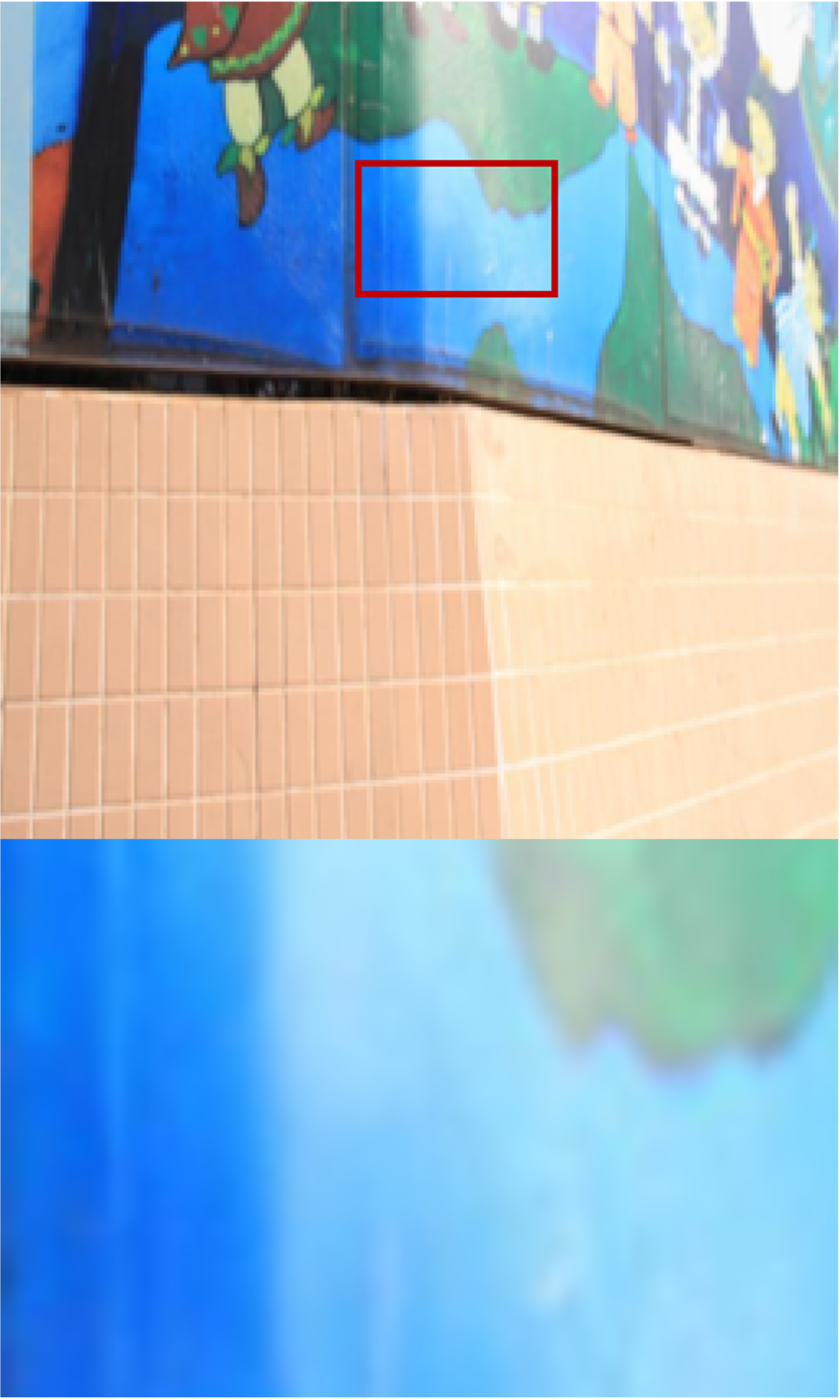}
    }\hspace{-6.5pt}
    \subfigure[HomoFormer]{
        \includegraphics[width=0.18\linewidth]{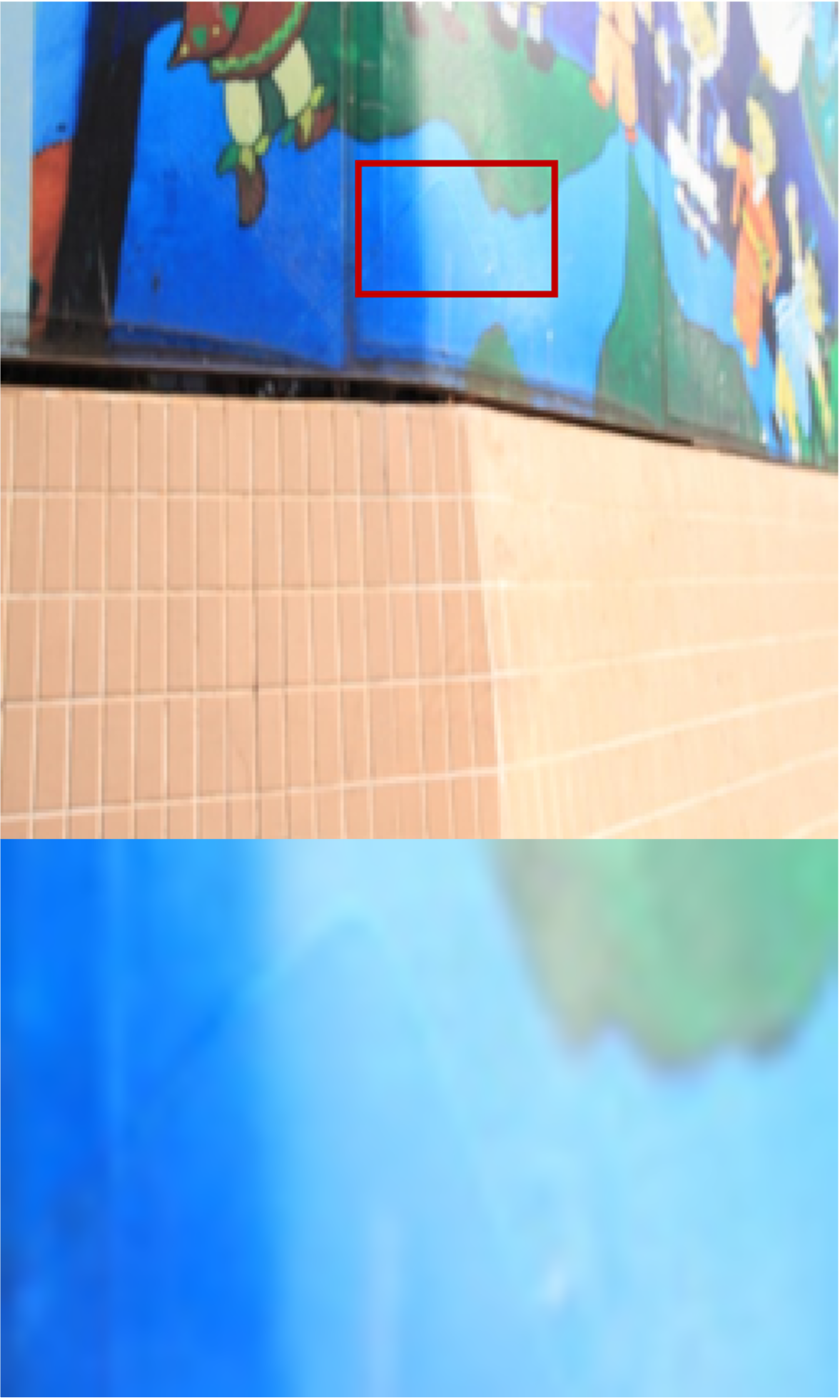}
    }\hspace{-6.5pt}
    \subfigure[Reference]{
        \includegraphics[width=0.18\linewidth]{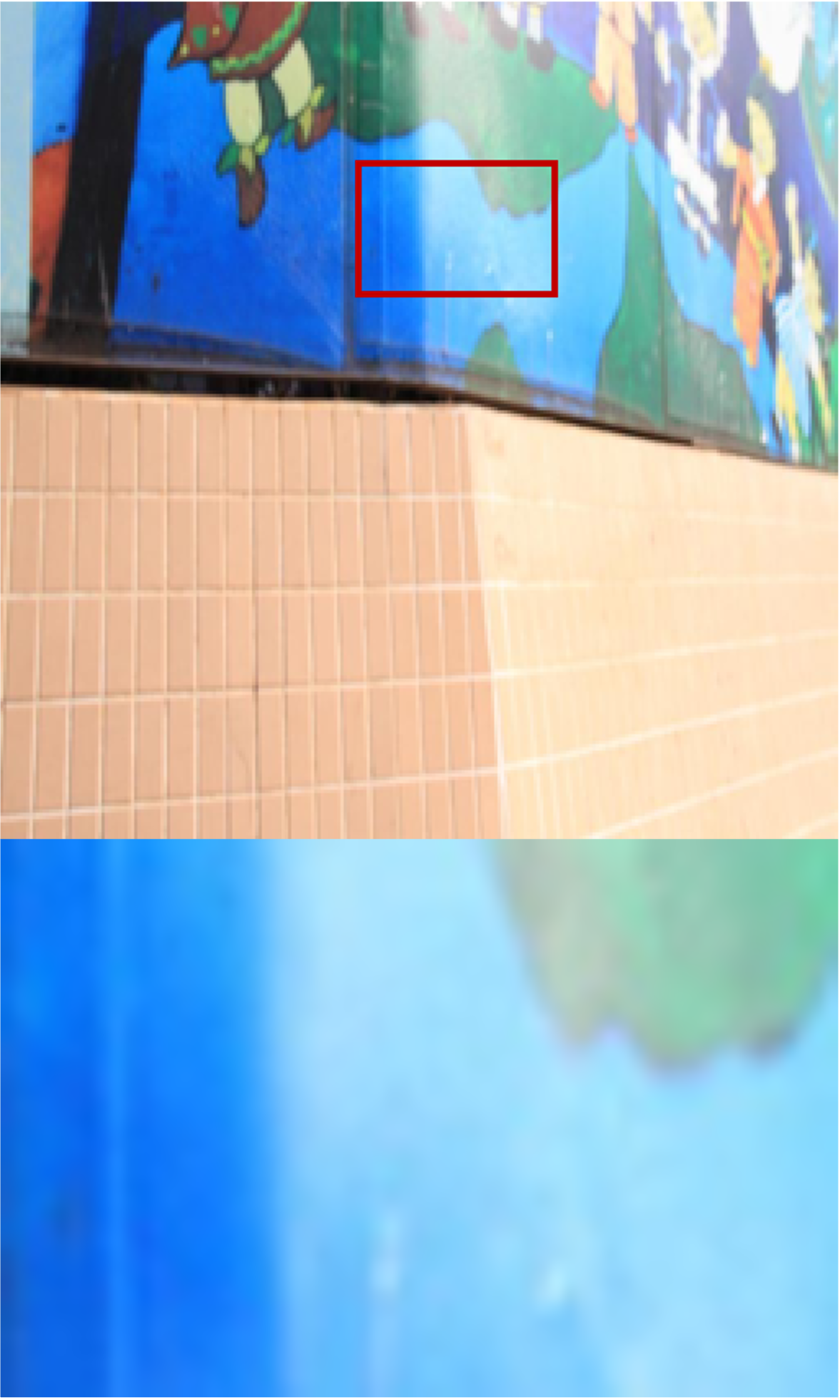}
    }
    \caption{Visual comparison with different shadow removal methods sampled from test set of SRD~\cite{qu2017deshadownet}.}
    \label{fig:srd2}
\end{figure*}

Tables~\ref{Table:srd}\&\ref{Table:istd+} show the quantitative results on the testsets over ISTD+, and SRD, respectively.
When the distribution of testing data is very similar to that of training data, supervised methods achieve better performance.
Typically, some recent transformer or diffusion based methods achieved state-of-the-art over ISTD+ and SRD datasets.
To further demonstrate the advantages and disadvantages of different methods, Figures~\ref{fig:srd}\&\ref{fig:srd2} present the visual examples of the shadow removal results on SRD dataset.
For hard shadow removal, the results often exhibit noticeable boundary artifacts due to the difficulty in distinguishing sharp shadow edges from structural edges in the original scene, as illustrated in the lower right corner of Figure~\ref{fig:srd}. 
It also highlights that in complex scenarios involving multiple shadow types, such as self-shadows and cast shadows, certain shadow regions are often overlooked by shadow detectors. This oversight results in significant residual shadows evident in the outputs of most methods.
Inaccurate masks can readily yield incorrect results across all types of shadows.
Existing methods have utilized a range of techniques, including joint mask refinement~\cite{guo2022shadowdiffusion}, regional classification~\cite{jin2022des3}, shadow matte prediction~\cite{le2019shadow}, to tackle the issue of mask inaccuracy, which enable better processing of ignored regions.

\subsection{Generalization Capability}
To further assess the generalization capabilities of current methods, we utilize the LRSS and UIUC datasets as our evaluation sets. These datasets are specifically chosen because they lack large-scale paired training data, enabling us to rigorously test how well the methods perform in scenarios where there is a distinct distribution difference between training and testing data.
For the LRSS dataset, we binarize the residuals of shadow-free images and shadow images to generate mask inputs for methods that require a shadow mask. For the UIUC dataset, we utilize the ground truth masks provided by the dataset.
To ensure a fair comparison, we consistently use the ISTD+ training set to train all methods.
The performances of most shadow removal methods significantly deteriorate when transferring to testing data that differs markedly from the training data. 
For instance, models trained on pairs from ISTD+ encounter difficulties in effectively handling shadow images from the soft shadow dataset LRSS, as illustrated in Table~\ref{Table:lrss_uiuc}. This is because the ISTD+ dataset predominantly consists of hard shadow types and lacks soft shadow and self-shadow samples.
On the contrary, those unsupervised or zero-shot learning methods may have better generalization ability to the unseen scenarios.
Some methods, like DC-ShadowNet, considered the classification of shadow types can have a better robustness cross various shadow types.
ShadowDiffusion employed the mask refinement as the auxiliary branch to achieve a better robustness to soft shadow processing.

\subsection{Computational Complexity}
In order to provide a comprehensive assessment of deep learning-based shadow removal techniques, we conduct a comparative analysis of the computational complexity of various state-of-the-art methods, as detailed in Table~\ref{tab:computational_cost}, including runtime, training parameters, and FLOPs over 100 images of size $256\times 256$ using one NVIDIA RTX A5000 GPU. 
Inpaint4Shadow~\cite{li2023leveraging} and DC-ShadowNet~\cite{jin2021dc} have the shortest runtimes.
BMNet~\cite{zhu2022bijective} employs a highly efficient architecture based on lightweight invertible blocks, resulting in significantly fewer model parameters and lower FLOPs.
Some methods use multiple sub-networks to perform multi-stage processing, which inevitably requires multiples of the parameters and computational load, such as SP+M-Net~\cite{le2019shadow}.
Specially, diffusion-based methods, such as ShadowDiffusion~\cite{guo2022shadowdiffusion} and BCDiff~\cite{guo2023boundary}, require iterative denoising steps.
Although multi-stage methods can achieve great performance, they require significantly increased computational complexity and processing time, making them impractical for many applications.

\begin{table*}[t]
 \renewcommand\arraystretch{1}
	\caption{Quantitative comparisons of \bihan{the algorithms' computational efficiency} in terms of runtime (in second), number of trainable parameters (\#Parameters) (in M), and FLOPs (in G).}
	\begin{center}
 \adjustbox{width=.85\linewidth}{
		\begin{tabular}{c|l|c|c|c|c}
			\hline
			\textbf{Learning}&	\textbf{Method}  
			& \textbf{RunTime}$\downarrow$ & \textbf{\#Parameters} $\downarrow$ & \textbf{FLOPs}$\downarrow$ & \textbf{Platform}\\
			\hline
			&	SP+M-Net \cite{le2019shadow}  & 0.023 &141.18 &61.52 &PyTorch\\
			&	DHAN \cite{cun2020towards} & 0.283 & 21.75 & 262.87 & TensorFlow\\
   			&	EMDN \cite{zhu2022efficient} & 0.015 & {\cellcolor{yellow!25}{10.06}} & 56.29  &PyTorch\\
      		&	BMNet \cite{zhu2022bijective} & 0.036 &{\cellcolor{red!25}{0.37}}&  {\cellcolor{red!25}{10.99}} &PyTorch\\
              		SL&	ShadowFormer \cite{guo2023shadowformer} & 0.031 & 11.35 & 64.60 & PyTorch\\
                & Inpaint4Shadow \cite{li2023leveraging} & {\cellcolor{red!25}{0.006}} & 14.98 & 81.18 & PyTorch\\
                		&	ShadowDiffusion \cite{guo2022shadowdiffusion} & 0.723  & 60.74  & 937.15  &PyTorch\\
                  		&	HomoFormer \cite{Xiao_2024_CVPR} & 0.042 & 17.81& {\cellcolor[HTML]{CCECEB}{35.63}} &PyTorch\\
			\hline
			UL&	DC-ShadowNet \cite{jin2021dc} & {\cellcolor{yellow!25}{0.008}} & {\cellcolor[HTML]{CCECEB}{10.59}} & {\cellcolor{yellow!25}{52.55}}  &PyTorch\\
			\hline
			\multirow{2}*{ZSL}	&	G2R-ShadowNet~\cite{liu2021from} & {\cellcolor[HTML]{CCECEB}{0.010}} & 22.76 & 113.93 &PyTorch\\			
			&	BCDiff \cite{guo2023boundary} &31.446 &276.41 & 139218.75 &PyTorch\\								
			\hline
		\end{tabular}}
	\end{center}
	\label{tab:computational_cost}
\end{table*}






\section{Shadow Removal Applications}\label{sec:application}

This section provides an overview of applications related to shadow removal and discusses its necessity, including shadow generation and shadow-related attacks.

\subsection{Shadow Generation}
Shadow generation and synthesis typically involve the process of creating or simulating shadows in digital images or computer graphics. Similarly to shadow removal, shadow generation serves as an important application, which can primarily be divided into the following two aspects according to their subsequent applications:
(1) Incorporating a foreground object into a background image to create a composite image relies heavily on the presence of shadows. Shadows are pivotal in rendering scenes realistically as they convey depth, perspective, and illumination.
(2) Additionally, shadow generation can aid in simulating large-scale paired images that include both shadows and shadow-free counterparts. This process is achieved at a minimal expense, contributing significantly to data augmentation.


\begin{figure*}[tbp]
    \centering
    \subfigure[Input]{
        \includegraphics[width=0.19\linewidth, height=0.19\linewidth]{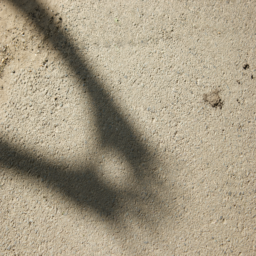}
    }\hspace{-6.5pt}
    \subfigure[Mask (Residual)]{ 
        \includegraphics[width=0.19\linewidth, height=0.19\linewidth]{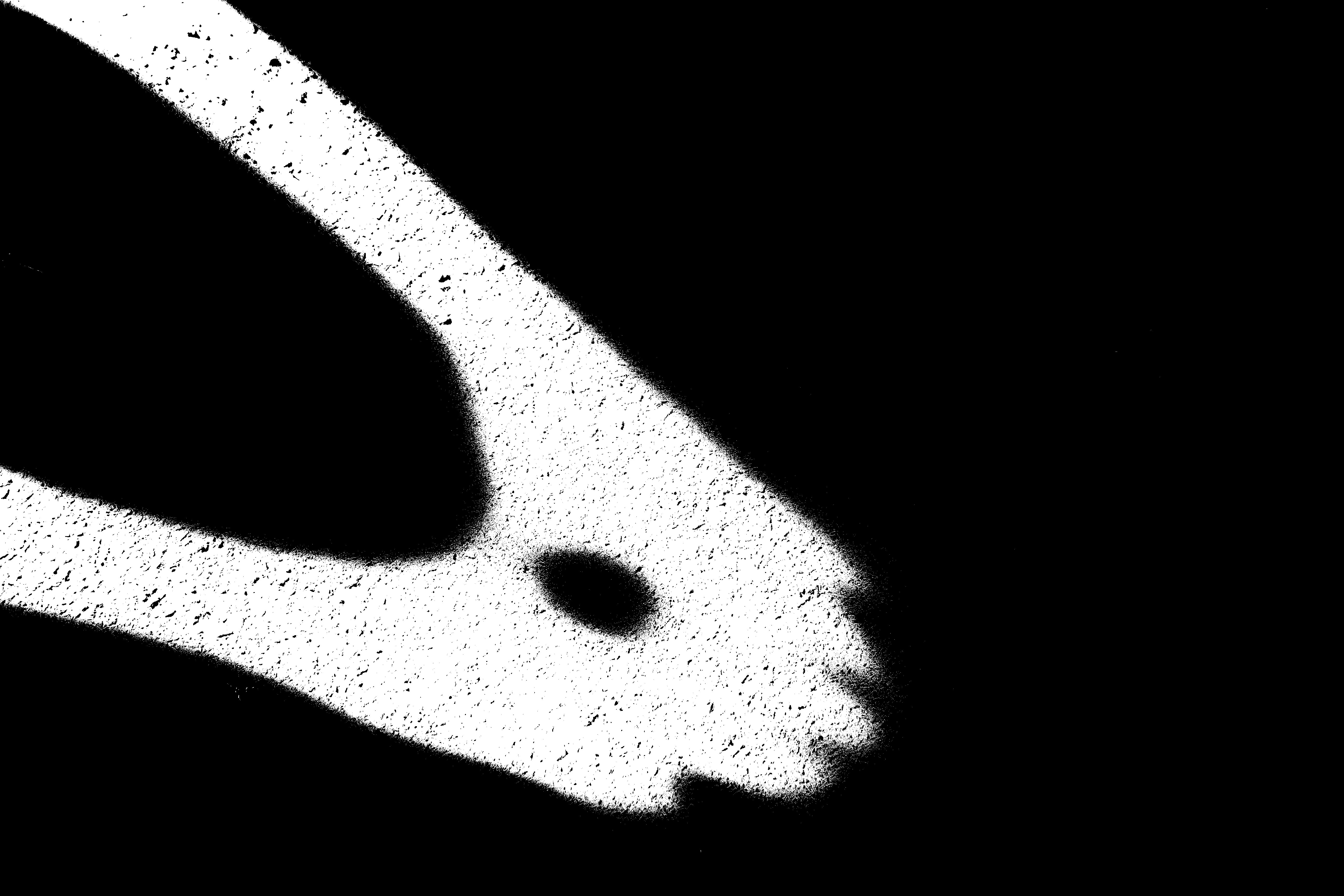}
    }\hspace{-6.5pt}
    \subfigure[SP+M-Net]{ 
        \includegraphics[width=0.19\linewidth, height=0.19\linewidth ]{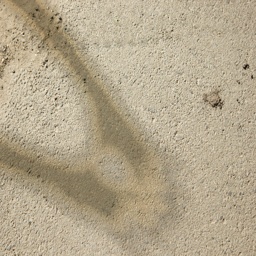}
    }\hspace{-6.5pt}
    \subfigure[EMDN]{ 
        \includegraphics[width=0.19\linewidth, height=0.19\linewidth]{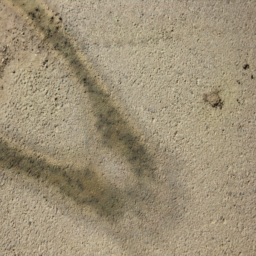}
    }\hspace{-6.5pt}
    \subfigure[G2R-ShadowNet]{ 
        \includegraphics[width=0.19\linewidth, height=0.19\linewidth]{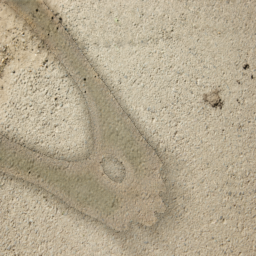}
    }
    \\
    \subfigure[Inpaint4Shadow]{ 
        \includegraphics[width=0.19\linewidth, height=0.19\linewidth]{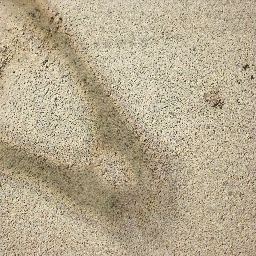}
    }\hspace{-6.5pt}
    \subfigure[ShadowFormer]{ 
        \includegraphics[width=0.19\linewidth, height=0.19\linewidth]{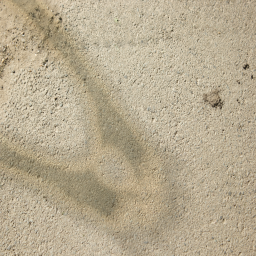}
    }\hspace{-6.5pt}
    \subfigure[ShadowDiffusion]{ 
        \includegraphics[width=0.19\linewidth, height=0.19\linewidth]{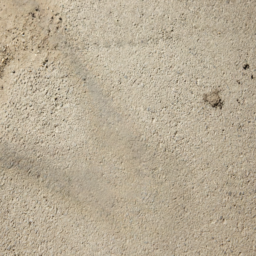}
    }\hspace{-6.5pt}
    \subfigure[HomoFormer]{
        \includegraphics[width=0.19\linewidth, height=0.19\linewidth]{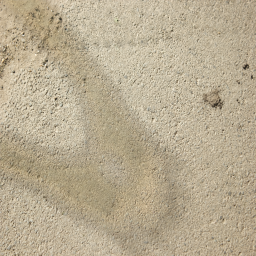}
    }\hspace{-6.5pt}
    \subfigure[GT]{
        \includegraphics[width=0.19\linewidth, height=0.19\linewidth]{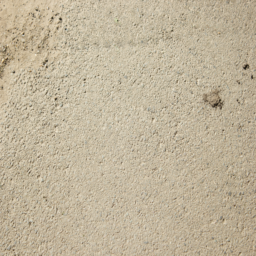}
    }
        \\
    \caption{Visual comparison with different shadow removal methods sampled from test set of LRSS~\cite{gryka2015learning}.}
    \label{fig:lrss}
\end{figure*}

\begin{figure*}[tbp]
    \centering
    \subfigure[Input]{
        \includegraphics[width=0.19\linewidth, height=0.19\linewidth]{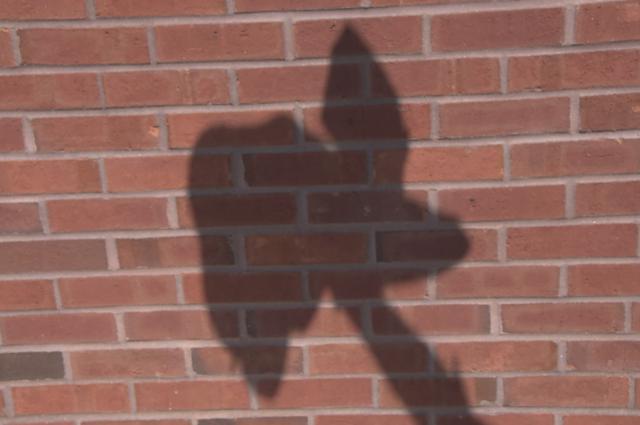}
    }\hspace{-6.5pt}
    \subfigure[Mask (Residual)]{ 
        \includegraphics[width=0.19\linewidth, height=0.19\linewidth]{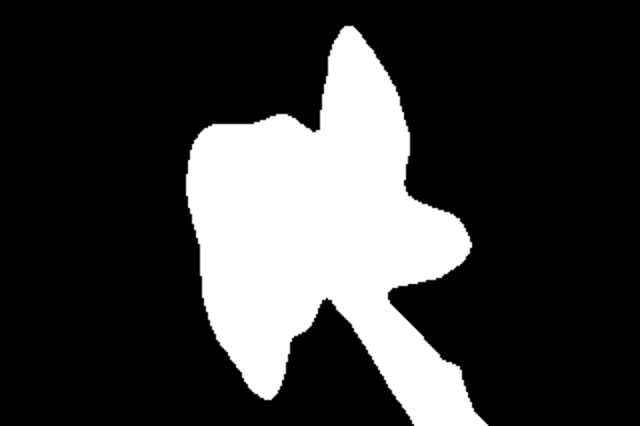}
    }\hspace{-6.5pt}
    \subfigure[SP+M-Net]{ 
        \includegraphics[width=0.19\linewidth, height=0.19\linewidth ]{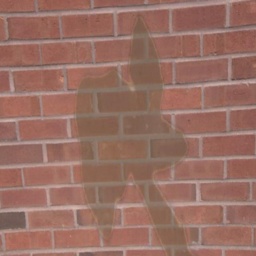}
    }\hspace{-6.5pt}
    \subfigure[EMDN]{ 
        \includegraphics[width=0.19\linewidth, height=0.19\linewidth]{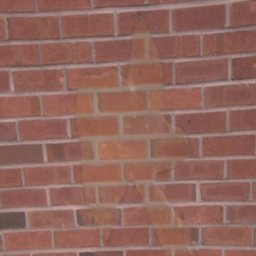}
    }\hspace{-6.5pt}
    \subfigure[G2R-ShadowNet]{ 
        \includegraphics[width=0.19\linewidth, height=0.19\linewidth]{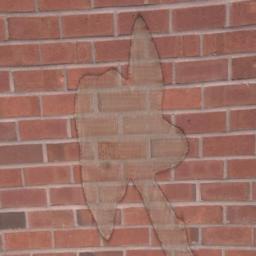}
    }
    \\
    \subfigure[Inpaint4Shadow]{ 
        \includegraphics[width=0.19\linewidth, height=0.19\linewidth]{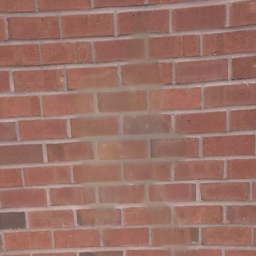}
    }\hspace{-6.5pt}
    \subfigure[ShadowFormer]{ 
        \includegraphics[width=0.19\linewidth, height=0.19\linewidth]{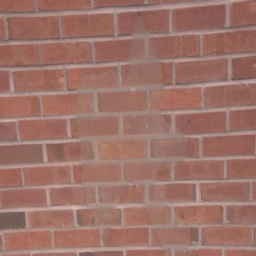}
    }\hspace{-6.5pt}
    \subfigure[ShadowDiffusion]{ 
        \includegraphics[width=0.19\linewidth, height=0.19\linewidth]{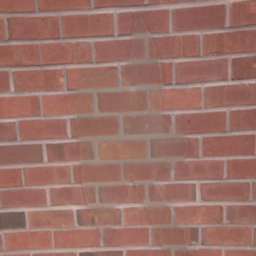}
    }\hspace{-6.5pt}
    \subfigure[HomoFormer]{
        \includegraphics[width=0.19\linewidth, height=0.19\linewidth]{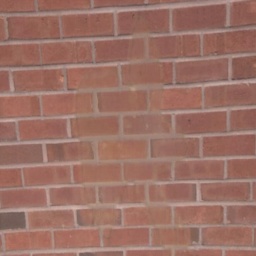}
    }\hspace{-6.5pt}
    \subfigure[GT]{
        \includegraphics[width=0.19\linewidth, height=0.19\linewidth]{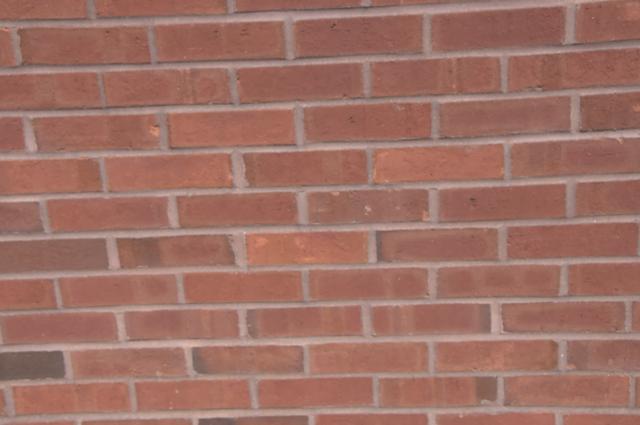}
    }
        \\
    \caption{Visual comparison \bihan{of the results using} different shadow removal methods, sampled from test set of UIUC~\cite{guo2012paired}.}
    \label{fig:uiuc}
\end{figure*}

\vspace{1mm}
\noindent\textbf{Image composition.}
Image composition involves inserting a foreground object into a background image to create a composite image. However, the quality of composite images can be greatly affected by inconsistencies between the foreground and background, including appearance, geometric, and semantic disparities. Shadow discrepancies represent a common issue in image composition, contributing to appearance inconsistencies. Therefore, shadow generation becomes essential to produce plausible shadows for foreground objects, enhancing the realism of the composite image.
Hong~\etal~\cite{hong2022shadow} introduced a manually annotated dataset alongside a shadow synthesis technique capable of predicting shadow masks. This method leverages illumination information from the background and employs an illumination model to estimate shadow parameters. However, manually removing shadows from scenes entails considerable cost.
Subsequently, Tao~\etal~\cite{tao2024shadow} proposed utilizing rendering techniques for 3D scenes to generate large-scale paired shadow/shadow-free images. They developed a two-stage network incorporating decomposed mask prediction and attentive shadow filling.


\vspace{1mm}
\noindent\textbf{Dataset augmentation.}
Due to the difficulty in obtaining extensive datasets for shadow removal, numerous methods in the field have opted to use the generation or synthesis process to supplement their training data.
Gryka~\etal~\cite{gryka2015learning} utilized graphics tools to produce shadow images, while Sidorov~\etal~\cite{sidorov2019conditional} constructed a shadow dataset by manipulating lighting within a video game. 
Nevertheless, computer-generated scenes and shadows often exhibit notable disparities from natural scenes~\cite{le2019shadow}.
To bridge this disparity, several methodologies~\cite{le2019shadow, hu2019mask, cun2020towards} have suggested employing Generative Adversarial Networks (GANs) to produce adversarial shadow images, aiming to enhance the realism and authenticity of synthesized shadows.
This approach leverages the adversarial training process to refine the quality and fidelity of synthesized shadows, ultimately enhancing their realism and alignment with natural scenes. 
Another category of methods (referenced in~\cite{inoue2020learning,guo2023boundary}) has utilized physical illumination models to simulate shadows with varying intensities, operating under the practical assumption that all occluding objects are situated outside the camera's field of view. 
This non-learning approach can be seamlessly integrated into existing training paradigms on-the-fly, offering a computationally inexpensive solution.





\subsection{Shadow-related Attacks}
With deep learning-based methods becoming the predominant solution for shadow removal, concerns regarding model-related robustness, such as adversarial robustness, have increasingly emerged within the community.

More recently, adversarial robustness in the physical domain has gained popularity due to its practical applications.
However, existing adversarial examples are generated in a ``sticker-pasting” strategy, which often falls short in producing imperceptible perturbation.
In contrast, shadow, as a ubiquitous natural phenomenon, has the potential to serve as a non-invasive adversarial perturbation.
Zhong~\etal~\cite{zhong2022shadows} propose a novel optical adversarial attack that generates naturalistic shadow patterns on real-world images under a black box setting.
As a consequence, they successfully demonstrated that shadow could achieve stealthy attacks that significantly misled the prediction of deep networks in both digital- and physical- settings.
Another line of research focuses on evaluating the robustness of existing deep shadow removal networks.
Due to the significant inconsistency of illumination between shadow and non-shadow regions, traditional adversarial attacks such as PGD~\cite{madry2018towards} fail to generate visually imperceptible noise, especially in the shadow region.
Motivated by this, Wang~\etal~\cite{wang2024benchmarking} propose a shadow-adaptive attack where the attack budget is adaptively aligned with each pixel's intensity.
Thus, the generated adversarial perturbation could be more aggressive in non-shadow regions while remaining stealthy in the shadow region.
With the merit of such shadow-adaptive attacks, they further establish the first benchmark evaluating the adversarial robustness of existing deep shadow removal networks.

\section{Future Research Directions}\label{sec:future}

As observed from the experimental results presented in Section~\ref{sec:benchmark}, shadow removal remains a challenging research topic, particularly concerning generalization ability. There remains considerable potential for improvement. This section suggests potential future research directions as follows:

\subsection{Generalized Shadow Removal}

\begin{enumerate}
    \item \textbf{From supervised to unsupervised/zero-shot learning.}
    While supervised learning-based shadow removal methods have demonstrated superior performance within specific datasets, their effectiveness diminishes significantly when applied to testing cases with distributions distinct from the training data.
    However, unsupervised/zero-shot learning methods offer a promising alternative. By harnessing the ability to capture the underlying structure and characteristics of data without necessitating labeled examples, these methods facilitate enhanced generalization to unseen, real-world scenarios.
    

    \item \textbf{Exploiting knowledge from pre-trained model.} 
Leveraging knowledge from pre-trained models, such as large language models~\cite{vaswani2017attention,kasneci2023chatgpt}, stable diffusion~\cite{sd2-1-base,sdxl}, and SAM (Spatially Adaptive Denoising Network)~\cite{kirillov2023segment}, holds immense potential for advancing shadow removal techniques. These models, having been trained on vast amounts of diverse data, encode rich representations of the underlying structure and semantics of the input data. By fine-tuning or incorporating these pre-trained models into shadow removal frameworks, researchers can tap into this wealth of knowledge to improve the quality and robustness of shadow removal algorithms.
    
\end{enumerate}

\subsection{Interactive Shadow Removal}


Interactive shadow removal~\cite{gong2014interactive,gong2016interactive} engages users in the shadow removal process through a user-friendly interface that allows real-time adjustments and feedback. This approach provides users with greater control and customization, enabling them to tailor the results to their specific preferences and needs. Although deep learning-based methods have shown promise, their application in this interactive domain remains less explored.

\begin{enumerate}
    \item \textbf{User-friendly input.} 
    Existing methods have overly stringent requirements for the input mask, or they rely on the off-the-shelf shadow detectors to generate the mask. However, using a pre-trained shadow detector may lead to a domain gap when applied to cases with different distributions from the training data, resulting in inaccurate shadow detection that could mislead the subsequent shadow removal network. Therefore, by utilizing simple user inputs such as clicks, strokes, or painted masks as prompts, the shadow removal network can effectively guide the process.

    \item \textbf{Real-time adjustment.} 
     The shadow removal process is optimized for real-time performance, ensuring that the results are generated quickly enough to be displayed without noticeable delay. This may involve leveraging hardware acceleration, parallel processing, or optimized algorithms to achieve fast computation speeds.
Additionally, in interactive shadow removal systems, users may provide feedback or make adjustments in real-time to refine the shadow removal results. 

    \item \textbf{Multiple shadows process.}
    In real-world scenarios, it's common to encounter multiple shadow regions, stemming from various sources such as multiple occluders or diverse light sources. When tackling these scenarios, users often seek the flexibility to selectively remove portions of these shadows based on user-defined input cues. This could involve removing shadows cast by specific objects or adjusting the intensity of shadows in certain areas, empowering users to tailor the shadow removal process to their specific needs and preferences.
    
\end{enumerate}

\subsection{More Comprehensive Benchmark}

Currently, the field of shadow removal faces a significant challenge due to the absence of a standardized evaluation benchmark. 
Existing test datasets for shadow removal may not adequately represent the diversity and complexity of real-world shadow scenarios. Some test data may be relatively straightforward to handle, whose distributions are distinct from real-world cases.
This discrepancy further complicates the evaluation process and undermines the reliability of comparative studies.
Hence, there is a clear need for a comprehensive benchmark that encompasses large-scale test samples, representing a diverse range of scenes including indoor and outdoor environments, as well as challenging lighting conditions such as various types of light sources and multiple shadows.
Establishing such a benchmark would enable researchers to conduct fair and rigorous evaluations of shadow removal methods, facilitating advancements in the field.




\subsection{Non-Reference Evaluation Metric}

Collecting paired shadow and shadow-free images poses a significant challenge in practical scenarios, mainly due to the complexities involved in capturing images under diverse lighting conditions. 
Furthermore, prevailing benchmarks heavily rely on reference metrics such as PSNR, RMSE, and SSIM for evaluation, which may not adequately account for scenarios where ground truth data or paired images are scarce or inaccessible.
In instances where large-scale shadow datasets featuring unpaired images, or those primarily collected for shadow detection purposes (\eg, SBU~\cite{hu2019mask} and USR~\cite{vicente2016large}), are available, leveraging them to the fullest extent for benchmarking becomes crucial. However, in such cases, the necessity for new non-reference metrics becomes apparent. These metrics offer a vital solution to the limitations posed by traditional evaluation methods, providing a more holistic and adaptable approach to image assessment. Introducing novel non-reference metrics addresses the shortcomings of traditional evaluation methods, offering a more comprehensive and adaptable approach to image assessment.





\subsection{Extension to Video Shadow Removal}
Current video shadow removal methods typically involve processing each frame of a video independently using image shadow removal algorithms, as mentioned by~\cite{liu2021from,guo2023boundary}. This approach, however, tends to neglect the temporal dimension of the video data. By treating each frame in isolation, these methods may fail to capture and exploit the temporal coherence present in video sequences.
When image-based algorithms are directly applied to videos, they often produce artifacts and inconsistencies across frames. This is because they do not account for the temporal context and variations in lighting conditions that occur over time. As a result, the shadow removal process may introduce flickering or other temporal artifacts, leading to unsatisfactory results.
Therefore, there remains a significant gap in the field of video shadow removal, calling for further research and development. 
Efforts are needed to explore novel algorithms that can effectively leverage the temporal dimension of video data. 


\subsection{Large-Scale Training Dataset}

While numerous training datasets are available for shadow removal, their scale and diversity in representing various shadow categories do not fully capture the complexity of real shadow degradations, as indicated in Table~\ref{tab:dataset}. Current deep shadow removal models, as discussed in Section~\ref{sec:benchmark}, encounter challenges in achieving satisfactory performance when confronted with shadow images captured in real-world scenarios. Moreover, the resolution of existing datasets is typically low, such as $480\times 640$, whereas real-world scenarios often involve high-resolution imagery, such as $2k$ or $4k$. Models trained on low-resolution data often struggle to generalize directly to high-resolution scenarios. Therefore, additional efforts are needed to explore the acquisition of large-scale, high-resolution, and diverse real-world paired shadow removal training datasets.

\section{Conclusion}\label{sec:conclusion}
In this paper, we present a comprehensive survey of deep learning-based single-image shadow removal approaches. We first discuss the formation of shadows in the real world and summarize the challenges of the shadow removal task. 
Building upon this, we explain how existing methods address these challenges from various perspectives, including training strategies, model architecture design, integration of physical models, and leveraging knowledge from detection models. We also conduct a comprehensive evaluation of existing methods with different learning strategies to explore their advantages and disadvantages. Based on the experimental results, we identify several open problems that still require further development, such as improving generalization to real-world cases, interactive shadow removal, more comprehensive benchmarks, video shadow removal, and large-scale training datasets.

\backmatter

\section*{Declarations}


\noindent \textbf{Conflict of interest.}  The authors declare that they have no known competing financial interests or personal relationships that could have appeared to influence the work reported in this paper.

\noindent \textbf{Availability of data and materials.} This work does not propose a new dataset. All the datasets we used are publicly available. 

\noindent \textbf{Code availability.} The related benchmark code and repository of this work is released at \url{https://github.com/GuoLanqing/Awesome-Shadow-Removal}.


\bibliographystyle{ieee_fullname} 
\bibliography{ref}
\end{document}